\pdfoutput=1
\documentclass[11pt]{article}
\usepackage{multirow}
\usepackage{booktabs}
\usepackage[table]{xcolor}
\usepackage[preprint]{acl}
\usepackage{amsmath}
\usepackage{times}
\usepackage{latexsym}
\usepackage{subcaption}
\usepackage{makecell}
\usepackage{adjustbox}
\usepackage{longtable}
\usepackage[T1]{fontenc}
\usepackage{booktabs}
\usepackage[utf8]{inputenc}

\usepackage{microtype}
\usepackage[table]{xcolor}
\usepackage{colortbl}
\usepackage{inconsolata}
\usepackage{tikz}
\usetikzlibrary{shapes, backgrounds}
\usepackage{graphicx}
\usepackage{changepage}
\definecolor{darkred}{rgb}{0.8,0.0,0.0}
\definecolor{lightred}{rgb}{1.0,0.7,0.7}
\definecolor{darkgreen}{rgb}{0.0,0.5,0.0}
\definecolor{lightgreen}{rgb}{0.7,1.0,0.7}
\definecolor{cherryred}{rgb}{0.8, 0.1, 0.1}
\definecolor{ocheryellow}{rgb}{0.8, 0.6, 0.2}

\newcommand{\CM}[1]{}
\newcommand{\new}[1]{\textcolor{black}{#1}}

\usepackage{cleveref}
\usepackage{comment}
\usepackage{todonotes}

\title{Child-Directed Language Does Not Consistently\\ Boost Syntax Learning in Language Models}

\author{
Francesca Padovani \quad
Jaap Jumelet \quad
Yevgen Matusevych \quad
Arianna Bisazza \\
Center for Language and Cognition (CLCG) \\ University of Groningen \\
\texttt{\{f.padovani, j.w.d.jumelet, yevgen.matusevych, a.bisazza\}@rug.nl}
}

\begin{document}
\maketitle

\begin{abstract}
Seminal work by \citet{huebner-etal-2021-babyberta} showed that language models (LMs) trained on English Child-Directed Language (CDL) can reach similar syntactic abilities as LMs trained on much larger amounts of adult-directed written text, %like Wikipedia, 
suggesting that CDL could provide more effective LM training material than the commonly used internet-crawled data.
However, the generalizability of these results across languages, model types, and evaluation settings remains unclear. 
We test this by comparing models trained on CDL vs.\ Wikipedia across two LM objectives (masked and causal), three languages (English, French, German), and three syntactic minimal-pair benchmarks. 
Our results on these benchmarks show inconsistent benefits of CDL, which in most cases is outperformed by Wikipedia models.
We then identify various shortcomings in previous benchmarks, and introduce a novel testing methodology, FIT-CLAMS, which uses a frequency-controlled design to enable balanced comparisons across training corpora.
Through minimal pair evaluations and regression analysis we show that training on CDL does \textit{not} yield stronger generalizations for acquiring syntax and highlight the importance of controlling for frequency effects when evaluating syntactic ability.\footnote{Code and data are available at \url{https://github.com/fpadovani/childes_vs_wiki}.}
\begin{comment}
\end{comment}
\end{abstract}

\section{Introduction}
The prevailing view in language acquisition research has long held that child-directed language (CDL) is inherently more effective than adult-directed language (ADL) for supporting first language %(L1) %AB acronym never used
development \cite{ferguson, schick:hal-03991945}. This has led to the assumption that the way caregivers speak to children is tailored to their developmental needs and functional for efficient language learning.

Motivated by this long-standing assumption, recent computational modeling research has investigated how training on CDL vs.\ ADL affects syntactic learning and generalization in neural network-based language models (LMs) 
\cite{feng-etal-2024-child,Mueller2023HowTP, yedetore-etal-2023-poor}.
Notably, \citet{huebner-etal-2021-babyberta} showed that BabyBERTa, a masked LM trained on 5M tokens of child-directed speech transcripts\footnote{Throughout this paper, we use the term CDL specifically to refer to transcripts of child-directed speech.}, achieves the level of syntactic ability similar to that of a much larger RoBERTa model trained on 30B tokens of ADL \cite{zhuang-etal-2021-robustly}.

Despite these encouraging findings, several issues complicate direct comparisons between CDL and ADL in LM training, including the variability in training setups \cite{cheng2023mcgill, feng-etal-2024-child, qin2024systematic} and evaluation benchmarks across studies \citep {warstadt2020blimp, huebner-etal-2021-babyberta, mueller-etal-2020-cross}, as well as the frequent focus on coarse-grained accuracy scores averaged over many syntactic paradigms.  
Moreover, recent work by \citet{kempe_ota_schaeffler_2024} reveals that the evidence for the facilitatory role of CDL in child language acquisition is scarce and specific to narrow domains, such as prosody and register discrimination, raising concerns about its generalizability. In this light, we believe it is crucial to carefully re-evaluate the benefits of CDL for LM training.

To better understand the specific effects of CDL as training input, we systematically compare LMs trained on CHILDES vs.\ Wikipedia across two architectures (RoBERTa and GPT-2) and three languages (English, French, and German) using four different benchmarks of minimal pairs. 
Crucially, we control for lexical frequency effects by introducing \textbf{FIT-CLAMS}, 
a Frequency-Informed Testing (FIT) methodology, which we apply to the CLAMS benchmark \cite{mueller-etal-2020-cross}.
The resulting evaluation set consists of minimal pairs balanced for subject and verb frequency in the training data, to disentangle true syntactic generalization from mere reliance on high-frequency lexical items present in the training data. We also perform a regression analysis to assess the impact of the distributional properties of CDL and ADL on the models' confidence in predicting grammaticality.

Our results challenge the presumed advantage of CDL for syntax learning in LMs, showing that it is, in fact, often outperformed by ADL. These findings underscore the need for a more nuanced understanding of when and how CDL may be beneficial, 
going beyond the mere adoption of CDL as training material---for instance,
%whether as a source of insights to improve training regimes in large-scale LMs, %(e.g., data augmentation with variation sets \cite{haga-etal-2024-babylm} or context variation \cite{xiao-etal-2023-towards}), or 
as a foundation to explore alternative learning paradigms that more closely mirror the interactive, contextual, and multimodal nature of human language acquisition \cite{beuls-van-eecke-2024-humans, stopler2025towards}.

\section{Related Work}

An ongoing debate in computational linguistics concerns whether CDL offers a measurable advantage over ADL in supporting the acquisition of formal linguistic knowledge in language models. The literature reports conflicting results: some highlight CDL’s benefits for grammatical learning and inductive bias, others find little or no advantage.
Among the studies supporting the benefits of CDL, a prominent example is \citet{huebner-etal-2021-babyberta}. Their study shows that LMs trained on CHILDES \cite{macwhinney2000childes}, a database containing transcripts of child--adult conversations, achieve higher average accuracy on Zorro, a minimal pair benchmark designed by \citet{huebner-etal-2021-babyberta}, compared to models trained on Wikipedia, when strictly controlling for dataset size and model configuration. An even better accuracy is achieved by LMs trained on written language adapted for children, such as AO-Newsela \cite{xu-etal-2015-problems}. 
\citet{salhan-etal-2024-less} report similar results in a cross-linguistic setting involving French, German, Chinese, and Japanese.
Across all four languages, their baseline RoBERTa-small model trained on CHILDES outperforms models trained on a size-matched Wikipedia corpus when evaluated on minimal-pair benchmarks available for each language. \citet{Mueller2023HowTP} further support the benefits of CDL by showing that pretraining LMs on simpler input promotes hierarchical generalization in question formation and passivization tasks, even with significantly less data than required by models trained on more complex sources like Wikipedia.
Finally, \citet{zora205878} leverage a non-contextualized word embedding model (Word2Vec by \citet*{word2vec}) to show that CDL is optimized for enabling semantic inference through lexical co-occurrence even in the absence of syntactic cues, suggesting that early meaning extraction in humans may be supported by surface-level regularities.

In contrast to studies highlighting the advantages of CDL, \citet{feng-etal-2024-child} report that models trained only on CDL underperform those trained on ADL datasets with higher structural variability and complexity (e.g., Wikipedia, OpenSubtitles, BabyLM Challenge dataset) in both syntactic tasks (like the ones in Zorro) and semantic tasks measuring word similarity. A similar result is reported by \citet{bunzeck2025construction}, who focus on German language models: while lexical learning tends to improve with the simpler, fragmentary language constructions typical of CDL, syntactic learning benefits from more structurally complex input. \citet{yedetore-etal-2023-poor} further challenge the benefits of CDL by demonstrating that both LSTMs and Transformers trained on CDL input fail to acquire hierarchical rules in yes/no question formation, instead relying on shallow linear generalizations. Finally, going beyond text-based LMs, \citet{gelderloos-etal-2020-learning} train their models on unsegmented speech data using a semantic grounding task and find that whilst child-directed speech may facilitate early learning, models trained on adult-directed speech ultimately generalize more effectively.

In light of such conflicting findings, our study offers a systematic reassessment of the impact of CDL on syntax learning
across two model architectures, three languages and multiple benchmarks, including a frequency-controlled one.

\section{Method}
We train RoBERTa- and GPT-2-style language models \textit{from scratch} on size-matched corpora of CHILDES and Wikipedia text in English, French and German. To assess their syntactic performance, we evaluate them on a set of existing minimal-pair benchmarks, enabling cross-linguistic and cross-architectural comparisons. Additionally, we propose FIT-CLAMS, a novel evaluation methodology inspired by \citet{mueller-etal-2020-cross}, which controls for lexical frequency effects and facilitates more reliable comparisons across datasets.

\subsection{Training Datasets}
We choose English for comparability to previous results, and French and German because they are included in the existing CLAMS benchmark \cite{mueller-etal-2020-cross}, enabling consistent cross-linguistic evaluation.\footnote{CLAMS also includes Hebrew and Russian, but these languages were not selected due to the limited amount of available CHILDES data.}
While typologically related, these languages provide sufficient variation, particularly in subject–verb agreement, to test the robustness of our findings.

We train models on two data types: CHILDES transcripts and Wikipedia.
For English, we use the same data split as \citet{huebner-etal-2021-babyberta}, comprising approximately 5M words of American English CDL, which in their work is referred to as AO-CHILDES \citep[Age-Ordered CHILDES;][]{huebner2021using} \footnote{In Table~\ref{tab:dataset-stats} we report 4.3M words, as our word count excludes punctuation and we have removed duplicated sentences from our validation split.}.
The French and German portions are extracted from CHILDES using the \texttt{childesr} library in R through the childes-db interface
\cite{sanchez2019childes}. 
We keep only adult-to-child utterances, excluding those produced by children. To enable fair comparisons, we sample Wikipedia corpora of matching sizes (measured in terms of whitespace-separated tokens). 
Despite the availability of other developmentally plausible, composite datasets such as the BabyLM Challenge \cite{warstadt2023proceedings} and the German BabyLM corpus \cite{bunzeck2025construction}, 
we exclusively use small-scale curated corpora from the CHILDES database to ensure a controlled and comparable experimental setup across languages, and to focus our evaluation on the interactive, infant-oriented register of CDL.

\begin{table}[t]
\centering
\small
\renewcommand{\arraystretch}{1.2}
\begin{adjustbox}{max width=\columnwidth}
\begin{tabular}{@{\ } l l @{\ \ } c @{\ \ }c @{\ \ }c @{\ \ }c @{\ \ } c @{\ }}
% \toprule
\multicolumn{2}{c}{} &
\multirow{2}{*}{\textbf{Tokens}} &
\textbf{Avg Sent.} &
\multicolumn{3}{c}{\textbf{Type/Token Ratio (TTR)}} \\
\arrayrulecolor[rgb]{0.753,0.753,0.753}\cline{5-7}\arrayrulecolor[rgb]{0.0, 0.0, 0.0}
\multicolumn{2}{c}{} & & \textbf{Length} & \textbf{1-grams} & \textbf{2-grams} & \textbf{3-grams} \\[0.9ex]
\midrule
\multirow{2}{*}{\textbf{EN}} 
  & CHILDES  & \multirow{2}{*}{ 4.3 M} & 6.13  & 0.005 & 0.073 & 0.275 \\
  & Wiki &                        & \new{24.89} & 0.026 & \new{0.289} & \new{0.682} \\
\arrayrulecolor[rgb]{0.753,0.753,0.753}\midrule\arrayrulecolor[rgb]{0.0, 0.0, 0.0}
\multirow{2}{*}{\textbf{FR}} 
  & CHILDES  & \multirow{2}{*}{2.3 M} & 6.48  & 0.009 & 0.089 & 0.310 \\
  & Wiki &                        & \new{38.58} & 0.022 & \new{0.190} & \new{0.468} \\
\arrayrulecolor[rgb]{0.753,0.753,0.753}\midrule\arrayrulecolor[rgb]{0.0, 0.0, 0.0}
\multirow{2}{*}{\textbf{DE}} 
  & CHILDES  & \multirow{2}{*}{3.8 M} & 5.61  & 0.012 & 0.129 & 0.424 \\
  & Wiki &                        & \new{21.95} & 0.056 & \new{0.385} & \new{0.757} \\
\bottomrule
\end{tabular}
\end{adjustbox}
\caption{Descriptive statistics of our training datasets.}
\label{tab:dataset-stats}
\end{table}

\begin{figure}[t]
     \centering
     \includegraphics[width=\linewidth]{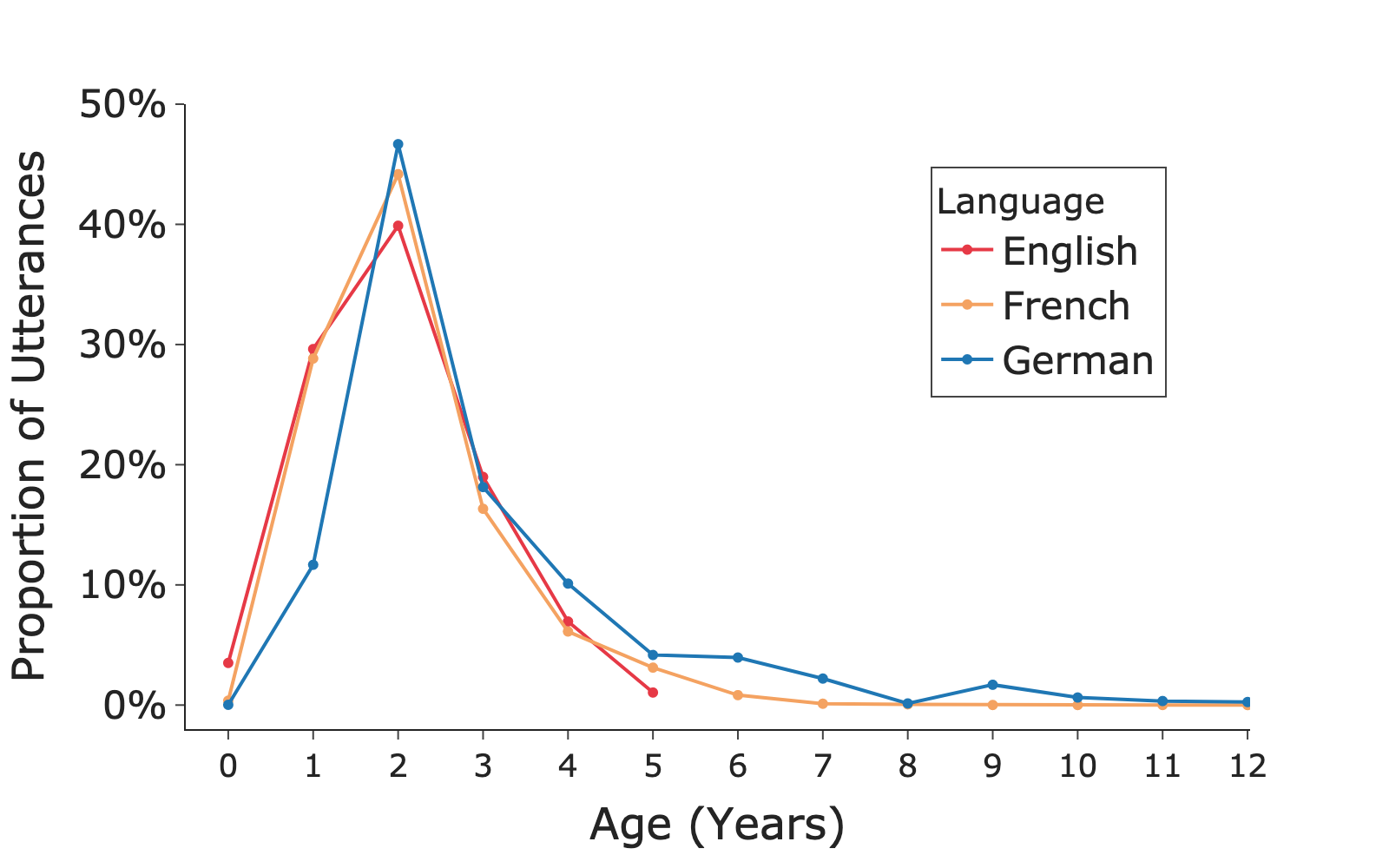}
    \caption{CHILDES age distribution across languages.}
     \label{fig:age_range}
 \end{figure}

Training data statistics are summarized in Table~\ref{tab:dataset-stats}. Across all languages, Wikipedia has substantially higher average sentence lengths compared to CHILDES. Additionally, notable disparities in type/token ratios reflect the highly repetitive nature of CDL, at both lexical and phrase level.
%further differentiate the two corpora. 
Further comparisons of the two corpora are presented in Appendix~\ref{sec:training_corpora}.
As for CHILDES-specific properties, Figure~\ref{fig:age_range} shows the data is heavily skewed toward the first 2–3 years of life, in all three languages.

\subsection{Models}

We evaluate two model architectures both based on Transformers \cite{vaswani}: \textbf{RoBERTa} \cite{zhuang-etal-2021-robustly}, a masked language model (MLM) chosen for consistency with prior CDL vs.\ ADL studies \cite{huebner-etal-2021-babyberta, salhan-etal-2024-less} and \textbf{GPT-2} \cite{radford2019language}, a causal language model (CLM), whose auto-regressive objective more closely approximates the incremental nature of human language processing \cite{goldstein2022shared}. To ensure a fair comparison, both models share the same architecture: 8 transformer layers, 8 attention heads, an embedding size of 512, and an intermediate feedforward size of 2048.

\subsubsection{Training Setup}
We train all models from scratch for 100,000 steps using the AdamW optimizer \cite{Loshchilov2017DecoupledWD} with linear scheduling and a warm-up phase of 40,000 steps.\footnote{During hyperparameter search, we experimented with various warm-up durations and found that shorter warm-up phases lead to early overfitting, particularly for GPT-2.} 
A learning rate of 0.0001, chosen for producing more stable learning curves, is applied consistently across all experiments. 
We use \new{92/8\%} train/validation split.

\section{Evaluation on Existing Benchmarks}

Following \citet{huebner-etal-2021-babyberta} and \citet{salhan-etal-2024-less}, we train separate Byte-Pair Encoding (BPE) tokenizers \cite{sennrich-etal-2016-neural} for each language and dataset, resulting in distinct vocabularies.\footnote{\citet{bunzeck2025subword} propose character-level models as a viable alternative for syntax learning, which could be tested in future CDL vs. ADL comparisons.} A vocabulary size of 8,192 tokens is used throughout, following prior developmental studies \cite{biemiller2003vocabulary} and consistent with earlier work in this area \cite{salhan-etal-2024-less}. Specifically, research has estimated that the average English-speaking 6-year-old has acquired approximately 5,000–6,000 words \cite{biemiller2003vocabulary}. Although our German and French CHILDES datasets are not strictly limited to children up to age six, the majority of the data comes from younger children, with relatively fewer samples from older age groups.

\subsection{Evaluation Procedure}

For evaluation, we report metrics averaged over three random seeds per model configuration. For MLMs, where no overfitting is observed, we use the final checkpoint at step 100,000. For CLMs, where overfitting \textit{is} observed, we select the checkpoint that yields the lowest validation perplexity for each language and training dataset.
Full validation perplexities trajectories are provided in Appendix~\ref{sec:models}, along with the list of selected checkpoints for each model and dataset (Table~\ref{tab:best-checkpoints}).

We assess the syntactic performance of a model by testing whether it assigns a higher probability to the grammatical version in a minimal sentence pair, 
a well-established paradigm in LM evaluation 
\cite{linzen-etal-2016-assessing, marvin-linzen-2018-targeted, wilcox-etal-2018-rnn}. Sentence probabilities are computed using the \texttt{minicons} library \cite{DBLP:journals/corr/abs-2203-13112}. For CLMs, we use the summed sequence log-probability with BOW correction.\footnote{Beginning-of-word (BOW) correction adjusts LM scoring by shifting the probability mass of `ending' a word from the BOW of the next token to the current one \cite{pimentel-meister-2024-compute,oh-schuler-2024-leading}.} For MLMs, we use the likelihood score with a within-word left-to-right masking strategy, which mitigates overestimation of token probabilities in multi-token words \cite{kauf-ivanova-2023-better}.

\subsection{Benchmark Description}

Several minimal-pair benchmarks have been developed to evaluate grammatical learning in models trained on CDL and ADL, most notably BLiMP \cite{warstadt2020blimp}, Zorro \cite{huebner-etal-2021-babyberta}, and CLAMS \cite{mueller-etal-2020-cross}. 

\begin{table*}[t]
\centering
\small
\renewcommand{\arraystretch}{1.2}
% \begin{adjustbox}{max width=\columnwidth}
\begin{tabular}{l l c c c c c}
 & & & & \multicolumn{3}{c}{\textbf{CLAMS}}\\
 \arrayrulecolor[rgb]{0.753,0.753,0.753}\cline{5-7}\arrayrulecolor[rgb]{0.0, 0.0, 0.0}
\textbf{Model} & \textbf{Training Data} & \textbf{BLiMP} & \textbf{Zorro} & {English} & {French} & {German}\\
\midrule
\multirow{2}{*}{\textbf{CLM}} 
  & CHILDES  & 0.61 $\pm$ 0.02 & \textbf{0.76 $\pm$ 0.04} & 0.60 $\pm$ 0.01 & 0.64 $\pm$ 0.01 & 0.69 $\pm$ 0.03 \\
  & Wiki & \new{0.61 $\pm$ 0.02}  & \new{0.69 $\pm$ 0.04}    & \textbf{0.71 $\pm$ \new{0.01}} & \textbf{\new{0.80} $\pm$ 0.01} & \textbf{0.81 $\pm$ 0.01}      \\
\arrayrulecolor[rgb]{0.753,0.753,0.753}\midrule\arrayrulecolor[rgb]{0.0, 0.0, 0.0}
\multirow{2}{*}{\textbf{MLM}} 
  & CHILDES  & 0.59 $\pm$ 0.03          & {0.66 $\pm$ 0.05} & 0.57 $\pm$ 0.02 & 0.59 $\pm$ 0.02 & 0.70 $\pm$ 0.01 \\
  & Wiki & 0.59 $\pm$ \new{0.03}         & \new{0.67 $\pm$ 0.03}  & \textbf{\new{0.63} $\pm$ 0.01} & \textbf{0.69 $\pm$ 0.01} & \textbf{\new{0.75} $\pm$ 0.01}         \\
\bottomrule
\end{tabular}
\caption{Model accuracies on BLiMP, Zorro, and CLAMS, averaged across paradigms and model seeds.}
\label{tab:results}
\end{table*}

\textbf{BLiMP} has become the standard benchmark for English, consisting of 67 paradigms representing 12 different linguistic phenomena. It is generated through a semi-automated process where lexical items are systematically varied within manually crafted sentence templates. While carefully controlled, this approach still produces semantically odd or implausible sentences \cite{martinez2023evaluating}. Moreover, this benchmark does not account for the vocabulary typical of CDL.

To address this lexical mismatch, \citet{huebner-etal-2021-babyberta} introduce
\textbf{Zorro}, a benchmark comprising 23 grammatical paradigms across 13 phenomena. 
Lexical items in Zorro's minimal pairs are selected by manually identifying entire words (never words split into multiple subwords) from the BabyBERTa subword tokenizer's vocabulary\footnote{The BabyBERTa tokenizer is jointly trained on AO-CHILDES, AO-Newsela, and an equally sized portion of Wikipedia-1.} and by counterbalancing word frequency distributions across the three training corpora. 
While this design enhances lexical compatibility across CDL and ADL training domains, selecting only non-segmented words overlooks the fact that models with robust syntactic understanding should be able to handle structure even when key items are split into subword units, raising concerns about the fairness and broader applicability of this benchmark.

Finally, \textbf{CLAMS} extends minimal pair coverage to five languages to enable a cross-lingually comparable syntactic evaluation, but only focuses on the phenomenon of subject–verb agreement (divided into 7 paradigms).
Despite its more limited syntactic scope compared to BLiMP and Zorro, we select CLAMS for our extended analyses, as subject-verb agreement represents a foundational aspect of grammatical ability which is typically acquired early in child language development \cite{b1464f919a1340418b46449ed6ef6461, 5GrammaticalIllusionsandSelectiveFallibilityinRealTimeLanguageComprehension}.
CLAMS is based on translations of minimal pairs originally created for English by \citet{marvin-linzen-2018-targeted}.
As stated by these authors, their models showed varied accuracy across specific verbs in the minimal pairs, with frequent ones like \textit{is} reaching 100\% accuracy and rarer ones like \textit{swims} only around 60\%, likely reflecting frequency effects. To account for such effects and ensure cross-linguistic consistency, we introduce in Section~\ref{fit_clams} a new methodology for constructing minimal pairs inspired by CLAMS, explicitly controlling for both verb and noun frequency across all language conditions.

\subsection{Results}
\label{sec:results_overall}

As shown in Table~\ref{tab:results}, the results obtained with our two model architectures are partially consistent with prior findings reported for English \cite{huebner-etal-2021-babyberta}.
In our experiments, both the causal and masked language models trained on CHILDES and Wikipedia perform comparably, with no significant differences in accuracy when tested on BLiMP.
On Zorro, the CLM trained on CHILDES outperforms its Wikipedia-trained counterpart, replicating the findings of \citet{huebner-etal-2021-babyberta}.
For the MLM architecture, the CHILDES-trained model shows only a modest accuracy advantage, smaller than that reported by \citet{huebner-etal-2021-babyberta}.

A more fine-grained analysis of the paradigms is provided in Appendix~\ref{sec:eval_bench_exist}, where Table~\ref{tab:clm_results_zorro}--\ref{tab:mlm_results_zorro} indicate that the CDL advantage is partly driven by grammatical phenomena involving questions. We find that this effect is even more pronounced in CLMs than in MLMs.
The trend  reflects the prevalence of interrogatives in the CDL data (40\% in English, see Appendix~\ref{sec:training_corpora}), which may bias models toward better handling of question-related constructions, as already noted by \citet{huebner-etal-2021-babyberta}.
To validate our evaluation pipeline and contextualize our results, we also test the models trained and released by \citet{huebner-etal-2021-babyberta}; details of this analysis are provided in Appendix~\ref{sec:eval_bench_exist}.

Focusing on subject--verb agreement, CLAMS results for the three languages are overall consistent across the two model types, but contradict the findings reported in previous work \cite{salhan-etal-2024-less}. For English, French and German, neither CLM nor MLM demonstrates an advantage when trained on CHILDES compared to Wikipedia, as shown in Table~\ref{tab:results}.

In summary, results are mixed: models trained on CDL sometimes perform better and sometimes worse than those trained on ADL, depending on the evaluation benchmark.
We hypothesize that lexical frequencies may be an important confounder in this type of evaluation, and
%Given these mixed results, we 
in the next section we set out to design new minimal pairs that balance the distribution of nouns and verbs representative of each training corpus.
As CLMs generally demonstrate higher performance than MLMs, we only focus on CLMs in our subsequent analyses. 

\section{FIT-CLAMS}

\begin{table*}[t]
\centering
\small
\renewcommand{\arraystretch}{1.2}
\begin{adjustbox}{max width=\textwidth}
\begin{tabular}{@{\ } l l @{\ \ } c @{\ \ } c @{\ \ } c @{\ \ } c @{\ }}
% \toprule
\multicolumn{1}{c}{} & \textbf{Minimal Pair} & \textbf{\#(noun;C)} & \textbf{\#(noun;W)} & \textbf{\#(verb;C)} & \textbf{\#(verb;W)} \\
\midrule
\multirow{3}{*}{\textbf{CLAMS}} 
 & \textit{the pilot [smiles/*smile]} & 73 & 263 & 180 & 19 \\
  & \textit{the author next to the guard [laughs/*laugh]} & 2 & 673 & 143 & 14 \\
  & \textit{the surgeon that admires the guard [is/*are] young} & 3 & 60 & 81,392 & 59,143 \\
\arrayrulecolor[rgb]{0.753,0.753,0.753}\midrule\arrayrulecolor[rgb]{0.0, 0.0, 0.0}
\multirow{3}{*}{\textbf{FIT-CLAMS-C}} 
  & \textit{the [resident/*residents] awaits} & 6 & 304 & 2 & 12 \\
  & \textit{the [farmer/*farmers] next to the guards arrives} & 264 & 169 & 17 & 102 \\
  & \textit{the [daddy/*daddies] that hates the friends thinks} & 7,028 & 5 & 14,710 & 227 \\
\arrayrulecolor[rgb]{0.753,0.753,0.753}\midrule\arrayrulecolor[rgb]{0.0, 0.0, 0.0}
\multirow{3}{*}{\textbf{FIT-CLAMS-W}} 
  & \textit{the [picker/*pickers] exaggerates} & 16 & 2 & 2 & 6 \\
  & \textit{the [painter/*painters] in front of the waiter enjoys} & 4 & 161 & 55 & 93 \\
  & \textit{the [president/*presidents] that admires the speakers works} & 49 & 1,473 & 2,012 & 3,545 \\
\bottomrule
\end{tabular}
\end{adjustbox}
\caption{
Minimal pair examples for CLAMS and FIT-CLAMS, and the noun and verb frequency across CHILDES (C) and Wikipedia (W) in each dataset.
}
\label{tab:minimal-pair-freqs}
\end{table*}

\label{fit_clams}
When comparing two models (trained on different data sets) on a syntactic evaluation task, we must ensure that any differences in their performance do not stem from the evaluation data being more `aligned' with the training distribution of one model over the other.
To achieve this, we propose a new Frequency-Informed Testing (FIT) evaluation methodology based on CLAMS, through which we generate \textit{two} sets of minimal pairs, each guided by the lexical distribution of a training corpus, ensuring a spread of high- and low-frequency items. % across each corpus. 

\subsection{Data Creation}
Our data creation follows the following four steps:
\begin{enumerate}
    \item \textbf{Vocabulary selection}: We compute the intersection of the vocabularies (\textit{before} applying subword tokenization) from Wikipedia and CHILDES, then select lexical items ensuring that all word forms have been encountered by \textit{both} models during training. 
    
    \item \textbf{Candidate selection}: Using SpaCy \citep{honnibal2017spacy}, we select candidate subjects and verbs by ensuring they have the right part-of-speech and grammatical features. Specifically, we select only animate nouns and limit verbs to present-tense third-person forms in indicative mood.
    We only keep nouns and verbs that occur in the corpora in both singular and plural form.
    
    \item \textbf{Controlling frequency}: To control for lexical frequency effects, nouns and verbs are grouped into 10 frequency bins based on their occurrence in the training data. 
    Frequency binning is done using uniformly spaced bins at a logarithmic scale, to account for the Zipfian distribution of word frequencies. 
    The distribution of nouns and verbs across bins is shown in Appendix~\ref{sec:noun_verb_distrib_bins}.
    From each frequency bin, one noun and one verb are manually selected from both the CHILDES and Wikipedia distributions, ensuring semantic compatibility.

    \item \textbf{Minimal pair creation}: 
    The final minimal pairs are generated adhering to the syntactic templates used by \citet{mueller-etal-2020-cross}.
    We adopt a minimal-pair design in which the critical region---the verb---is held constant across grammatical and ungrammatical conditions, as is also done in BLiMP-NL \cite{blimpnl}.
    Evaluating model probabilities only at this critical region (while changing the context) avoids confounding effects from differences in subword tokenization.
    
\end{enumerate}

Following this pipeline, we generate two sets of minimal pairs, one from CHILDES distribution (FIT-CLAMS-C) and one from Wikipedia distribution (FIT-CLAMS-W), forming together the FIT-CLAMS benchmark. In total, we generate 16,400 minimal pairs for English, 4,914 for French, and 10,800 for German across the various paradigms (see Table~\ref{tab:minimal-pairs} for detailed paradigm counts). Minimal pair examples, together with the corresponding noun and verb frequencies in both CHILDES and Wikipedia data, are provided in~\Cref{tab:minimal-pair-freqs}.

\subsection{Results}

Our minimal pairs are constructed such that the verb remains constant across the pair. Consequently, we compute the model's probability for the verb alone, rather than over the entire sentence, as was done for the three previous benchmarks.
This approach probes more directly the model’s syntactic ability by correctly solving subject-verb agreement, assigning higher probability to the verb form that matches the sentence-initial subject. 
Each CLM (whether trained on CHILDES or Wikipedia) is evaluated on both sets. 
Depending on the lexical source, each evaluation set may be considered either in-distribution or out-of-distribution relative to a model's training data.

\begin{table}[t]
\centering
\renewcommand{\arraystretch}{1.2}
\begin{adjustbox}{max width=\columnwidth}
\begin{tabular}{@{\ } l @{\ \ \ } l c @{\ \ }c @{\ \ }c @{\ }}
% \toprule
\textbf{Training} & \textbf{Eval. lex.} & \textbf{EN} & \textbf{FR} & \textbf{DE} \\
\midrule
\multirow{2}{*}{CHILDES} 
  & CHILDES & {0.63 $\pm$ 0.02} & {0.78 $\pm$ 0.04} & {0.73 $\pm$ 0.03} \\
  & Wiki & 0.63 $\pm$ 0.03 & 0.67 $\pm$ 0.03 & 0.69 $\pm$ 0.04 \\
\arrayrulecolor[rgb]{0.753,0.753,0.753}\midrule\arrayrulecolor[rgb]{0,0,0}
\multirow{2}{*}{Wiki} 
  & CHILDES & \textbf{\new{0.72 $\pm$ 0.03}}  & \textbf{\new{0.86 $\pm$ 0.02}}  & \textbf{\new{0.83 $\pm$ 0.02}} \\
  & Wiki & \textbf{\new{0.75 $\pm$ 0.02}} & \textbf{\new{0.88 $\pm$ 0.06}} & \textbf{\new{0.82 $\pm$ 0.03}} \\
\bottomrule
\end{tabular}
\end{adjustbox}
\caption{Average accuracy of the CLM models evaluated on the two FIT-CLAMS versions based, respectively, on the lexical distribution of CHILDES and Wikipedia. Best scores per dataset are shown in boldface. 
}
\label{tab:new_clams_results_overall}
\end{table}

\begin{figure}[htbp]
    \centering
    \includegraphics[width=0.48\textwidth]{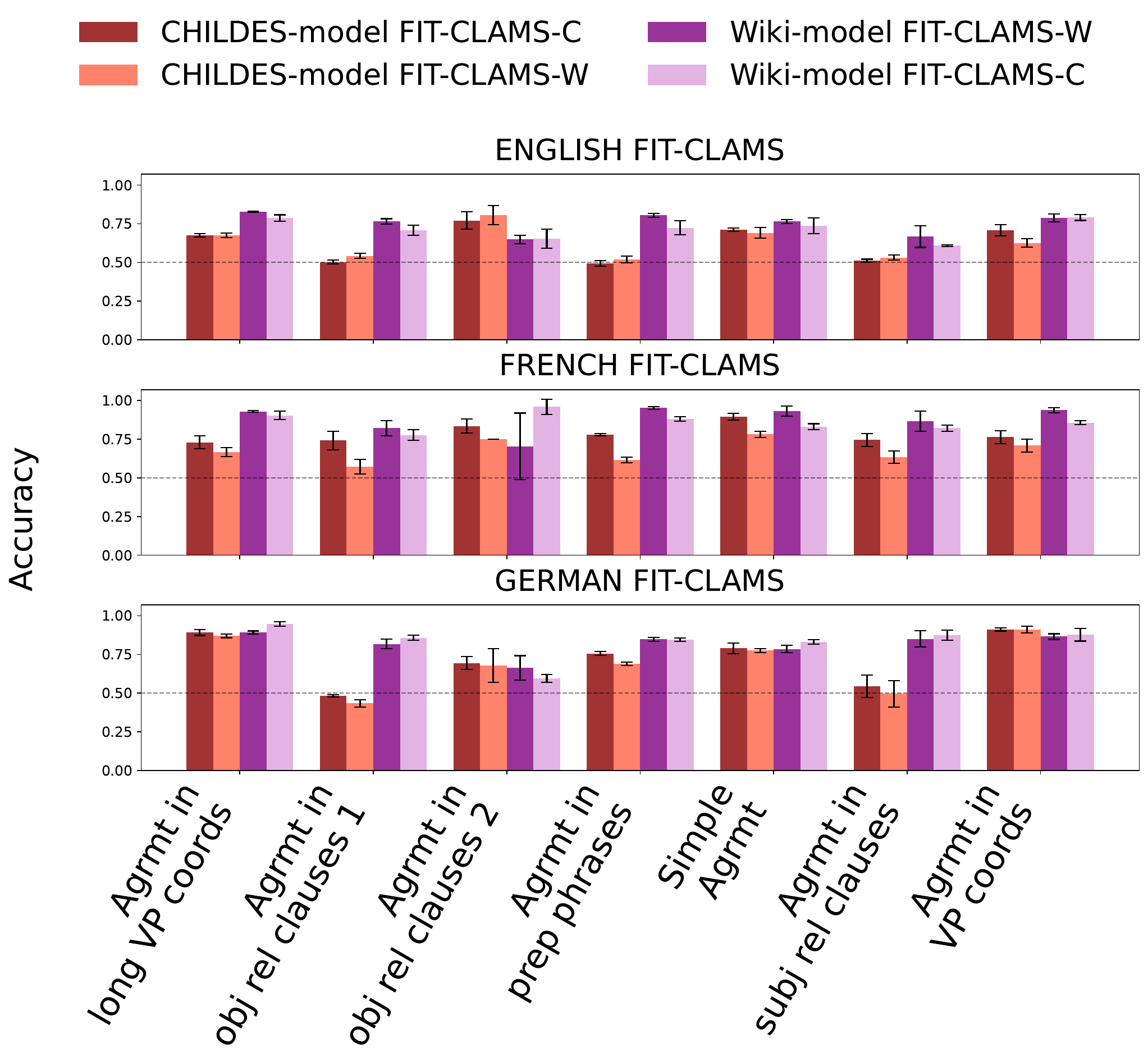}
    \caption{
    Accuracy of our models on the individual paradigms in the new set of minimal pairs, FIT-CLAMS.}
    \label{fig:fit-clams-results}
\end{figure}

Table~\ref{tab:new_clams_results_overall} presents average accuracy scores across the seven syntactic paradigms. 
Overall, we observe that average accuracy on FIT-CLAMS increases for both models (trained on CHILDES and Wikipedia) compared to their performance on CLAMS (see Table~\ref{tab:results}). This increase can be explained by the fact that in the original CLAMS dataset, some minimal pairs contain tokens that are not observed at training time, which is not the case for FIT-CLAMS. These results also reveal that, as expected, models generally perform better on minimal pairs constructed with in-distribution lexical items than with out-of-distribution ones (except the German model trained on Wikipedia).

Importantly, the most pronounced contrast is still the one between the models trained on CHILDES (first two rows) vs. Wikipedia (last two rows): the latter consistently outperform the former across all languages on the subject--verb agreement task. 
Thus, even when strictly controlling for lexical frequency, models trained on Wikipedia continue to show a systematic advantage, underscoring the benefits of training on larger and more diverse textual resources for developing robust syntactic abilities.

\section{Regression Analysis}
To further investigate how training data shapes model behavior, we conduct a linear regression analysis examining whether and how the presence of specific lexical items in the training data influences model performance. 
A model that builds up a robust representation of number agreement will be better able to generalize to infrequent constructions, without relying on memorization \citep{lakretz-etal-2019-emergence,patil-etal-2024-filtered}.
We focus on Simple Agreement, as in this paradigm it is straightforward to connect the frequency of occurrence of individual words in the training data to subsequent model performance. 
Specifically, we assess how unigram frequency of critical lexical items---the subject and the verb---affects the model’s preference for grammatical over ungrammatical sentences. 
This controlled setup allows us to isolate frequency effects and compare the degree to which models trained on CDL and ADL generalize beyond lexical co-occurrence patterns. 

We fit ordinary least squares (OLS) regressions to the training data (CHILDES or Wikipedia in three different languages) and the probabilities generated by LMs. 
Specifically, the \textit{dependent variable} used in the regression analysis is the \textit{$\Delta P$-score}, defined as the difference between the probability assigned by the model to the verb in a grammatical and ungrammatical context:
\[\Delta P(v|c) = \log P(v | c^+) - \log P(v | c^-)\]
for verb $v$ in a grammatical ($c^+$) and ungrammatical context ($c^-$): e.g., \textit{The boy walks} vs. $^*$\textit{The boys walks}.
As \textit{independent variables}, we use the log-frequencies of (1) the verb, (2) the grammatical subject noun, and (3) the ungrammatical subject noun, fitting a single multivariate model with all variables.
Lexical frequency values are extracted from the corpus used to train the respective model (either CHILDES or Wikipedia).
All predictor variables are standardized using z-score normalization. 

\begin{figure}
    \centering
    \includegraphics[width=\columnwidth]{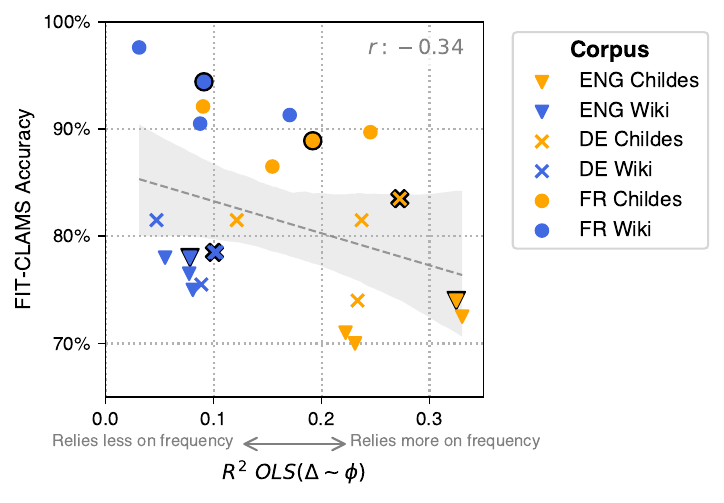}
    \caption{
    Relation between LM accuracy on FIT-CLAMS and proportion of variance ($R^2$) explained by the OLS regression fitted on lexical frequency factors.
    The lower the $R^2$ is, the less the LM's behavior is driven by lexical frequency.
    Each LM configuration is represented by four data points: three individual LMs (random seeds) and the average of the three \new{(highlighted with black outline)}.
    }
    \label{fig:r2_acc}
\end{figure}

To investigate the impact of lexical frequency, we examine the relationship between the fit (i.e., $R^2$) of the OLS regression and the LMs' accuracy on the FIT-CLAMS data.
Our hypothesis is that the predictions of an LM will be less driven by frequency if it generalizes well beyond the sentences it saw during training, and as such the OLS will lead to a \textit{lower} $R^2$ score. 
The results in Figure~\ref{fig:r2_acc} reveal a moderate negative correlation between $R^2$ and accuracy \new{($r=-0.34$, $p=0.10$)}: the best-performing LM (trained on French Wikipedia) yields the lowest $R^2$, whereas the worst-performing LM (trained on English CHILDES) yields the highest.
Although models trained on CHILDES data perform slightly better for German, they are nonetheless more influenced by lexical frequency compared to Wikipedia-trained models.
We hypothesize this is partly driven by the Type/Token Ratio (TTR), which is \new{higher for Wikipedia than CHILDES} (see Table~\ref{tab:dataset-stats}).
Generalization in LMs is driven by \textit{compression}: by being forced to build up representations for a wide range of inputs in a bounded representation space, models have to form abstractions that have been shown to align with linguistic concepts \citep{tishby2015, tenney-etal-2019-bert, wei-etal-2021-frequency}.
Training models on low-TTR data, therefore, leads to a weaker generalization than on high-TTR data, since a model trained on low-TTR data can rely more on memorization.
We leave a more detailed exploration of such factors to future work.

\section{Discussion and Conclusions}

This study examined whether language models trained on CDL can match or surpass the syntactic ability of models trained on size-matched ADL data. We trained RoBERTa- and GPT-2-based models in English, French, and German and evaluated them on multiple minimal-pair benchmarks as well as on the newly introduced, frequency-controlled FIT-CLAMS. The results show that CHILDES-trained models \textit{underperform} Wikipedia-trained models in most cases. % , across languages, architectures, and evaluation settings. 
Our regression analysis further reveals a modest negative correlation
between model accuracy and variables based on lexical frequency, indicating that stronger models rely less on surface-level patterns of lexical co-occurrence.

When interpreting these findings through the lens of language acquisition, it is important to consider the limitations of the training paradigm we use. Our models are trained in artificial conditions that diverge substantially from the way humans acquire language. Unlike children, these models are exposed to static datasets without any form of interaction, feedback, or communicative pressure. Additionally, the learning process is not incremental or developmentally grounded, the vocabulary is extracted from the entire corpus at once when the tokenizer is trained, and the models operate without cognitive constraints or working memory limitations. 
These discrepancies highlight an important gap between current computational learning frameworks and the dynamics of natural language acquisition.

Rather than completely dismissing CDL, we contend that it should be recontextualized and rigorously tested within frameworks that better resemble human language learning processes. CDL might hold particular promise when integrated into models that simulate interactive, situated communication \citep{beuls-van-eecke-2024-humans, stopler2025towards}, shifting the focus toward the communicative and contextual factors essential to language acquisition, which are absent in static text-based training regimes.
Moreover, LM experiments can still contribute significantly to the study of human language acquisition \cite{warstadt2022artificial,pannito,portelance2024roles}, where the benefits of CDL remain poorly understood \citep{kempe_ota_schaeffler_2024}, by helping to uncover specific properties of CDL that make it particularly suitable for specific kinds of learning outcomes. For instance, scaling up experiments like those of \citet{zora205878} could provide valuable insights into various aspects of language acquisition, such as morphological, syntactic, and semantic development.

Finally, rather than serving solely as pretraining data, CDL, together with insights from the language acquisition literature \cite{kempe_ota_schaeffler_2024}, can inspire the design of inductive biases and data augmentation strategies, such as context variation \cite{xiao-etal-2023-towards} or variation sets \cite{haga-etal-2024-babylm}, with the practical aim of improving generalization or enabling more data-efficient learning in models trained on the standard adult-directed text corpora that are used in NLP applications. 

In conclusion, although CDL does not appear to improve syntactic learning in conventionally trained LMs,
%although conventional training on CDL does not currently improve syntactic learning in LMs, 
we maintain that it remains a valuable resource deserving further investigation. Future work should prioritize CDL's integration within cognitively and interactively grounded frameworks, while also exploring how its distinctive characteristics can inform the development of more effective model architectures and training methodologies.

\section*{Limitations}
This work does not explicitly account for certain grammatical inconsistencies characteristic of child-directed language, such as the frequent use of infinitive verb forms in contexts where a third- or first-person singular subject is intended, resulting in subject-verb agreement violations.
Such errors, which have been systematically mapped for English in an extensive taxonomy by \citet{nikolaus-etal-2024-automatic}, may introduce noise into the expression of subject–verb agreement. We hypothesize that these properties of CDL could affect the grammatical learning of this syntactic phenomenon. Future experiments could explore whether removing or correcting these occurrences in the training data improves model performance on subject-verb agreement tasks.
Another limitation concerns the restricted syntactic scope of our regression analysis, which is limited to simple cases of subject–verb agreement. More structurally complex agreement configurations, such as those involving long-distance dependencies, coordination structures, or prepositional phrases, are not included in the current regressions. In future work, we plan to broaden the analysis to these more challenging constructions to examine whether surface-level factors like lexical frequency continue to influence model performance, and how these effects may differ between CHILDES- and Wikipedia-trained models.

The lexical diversity of our regression setup also imposes constraints on the generalizability of our findings. 
While FIT-CLAMS covers a considerably wider spectrum of lexical frequencies compared to the original CLAMS, each syntactic item (verb or noun) is represented by at most 10 lexical instances per language, with as few as 7 for French verbs. Expanding this set to include a broader and more representative frequency distribution would allow for more robust and precise estimates of how lexical frequency relates to syntactic generalization.

Finally, the construction of our FIT-CLAMS benchmark involved manual selection of animate nouns and semantically compatible verbs shared across CDL and Wikipedia corpora. While this ensured controlled and interpretable comparisons, it limits scalability. In future work, automating this process, could facilitate broader and more flexible evaluations.

\bibliography{main}

\appendix

\section{Training Corpora}
\label{sec:training_corpora}
\begin{figure*}[ht]
     \centering
     \includegraphics[width=\linewidth]{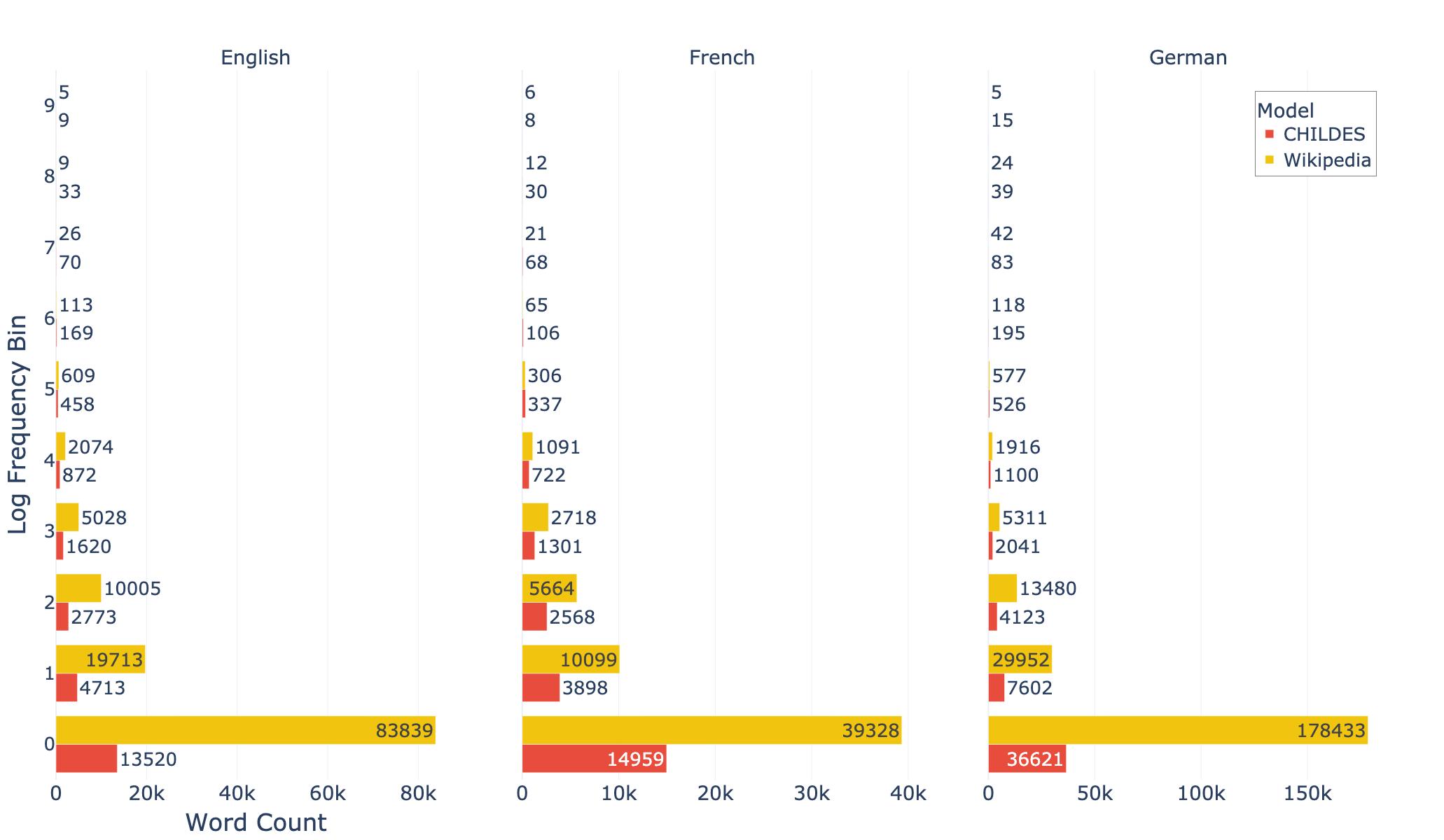}
    \caption{Word Frequency Distribution (CHILDES vs Wikipedia) across languages.}
     \label{fig:word_dist_lang}
 \end{figure*}

\begin{figure*}[h]
     \centering
     \includegraphics[width=\linewidth]{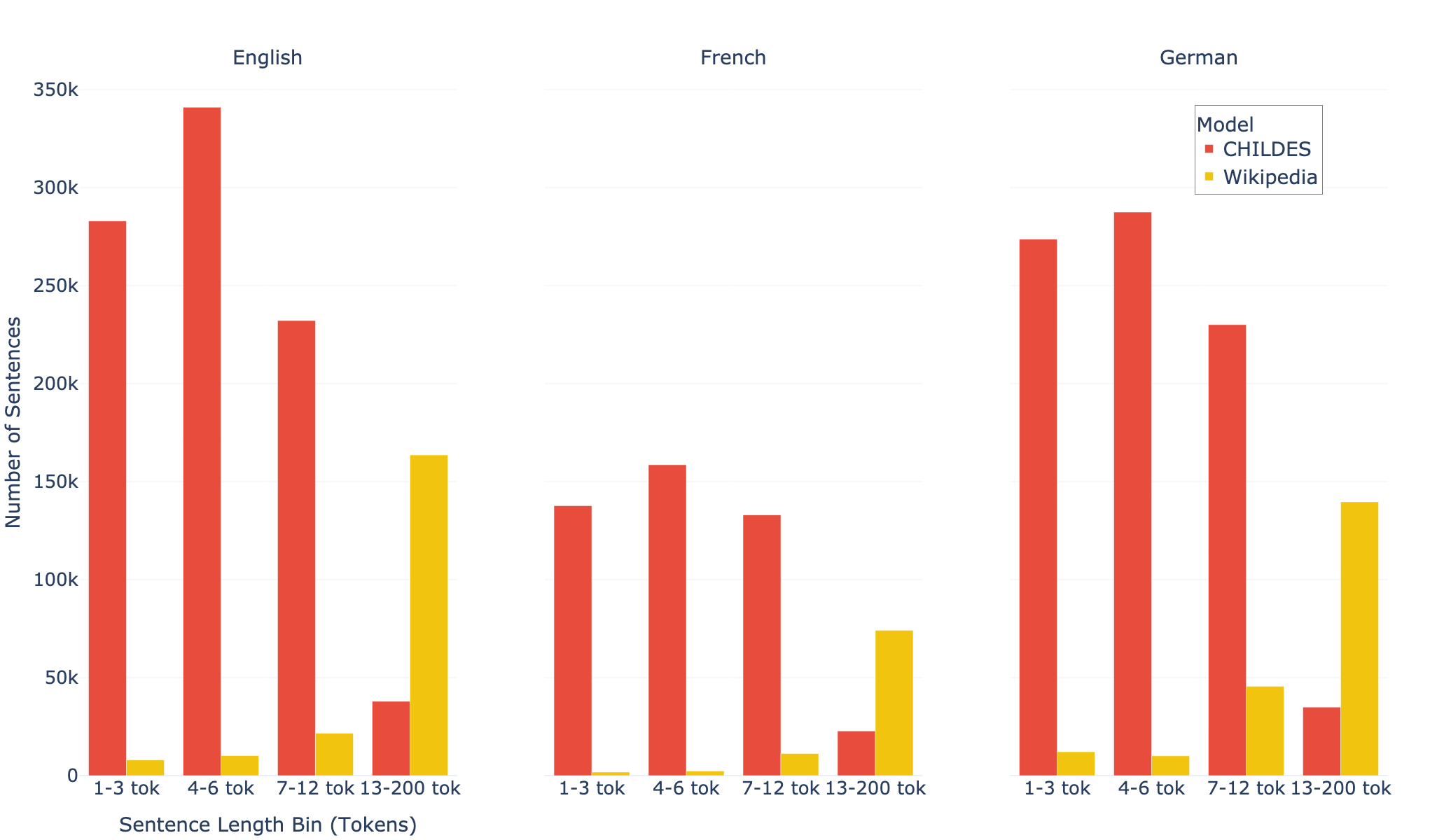}
    \caption{Sentence Length Distribution (CHILDES vs Wikipedia) across languages and data types}
     \label{fig:sentence_length_dist_lang}
 \end{figure*}

\begin{figure*}[h]
     \centering
     \includegraphics[width=\linewidth]{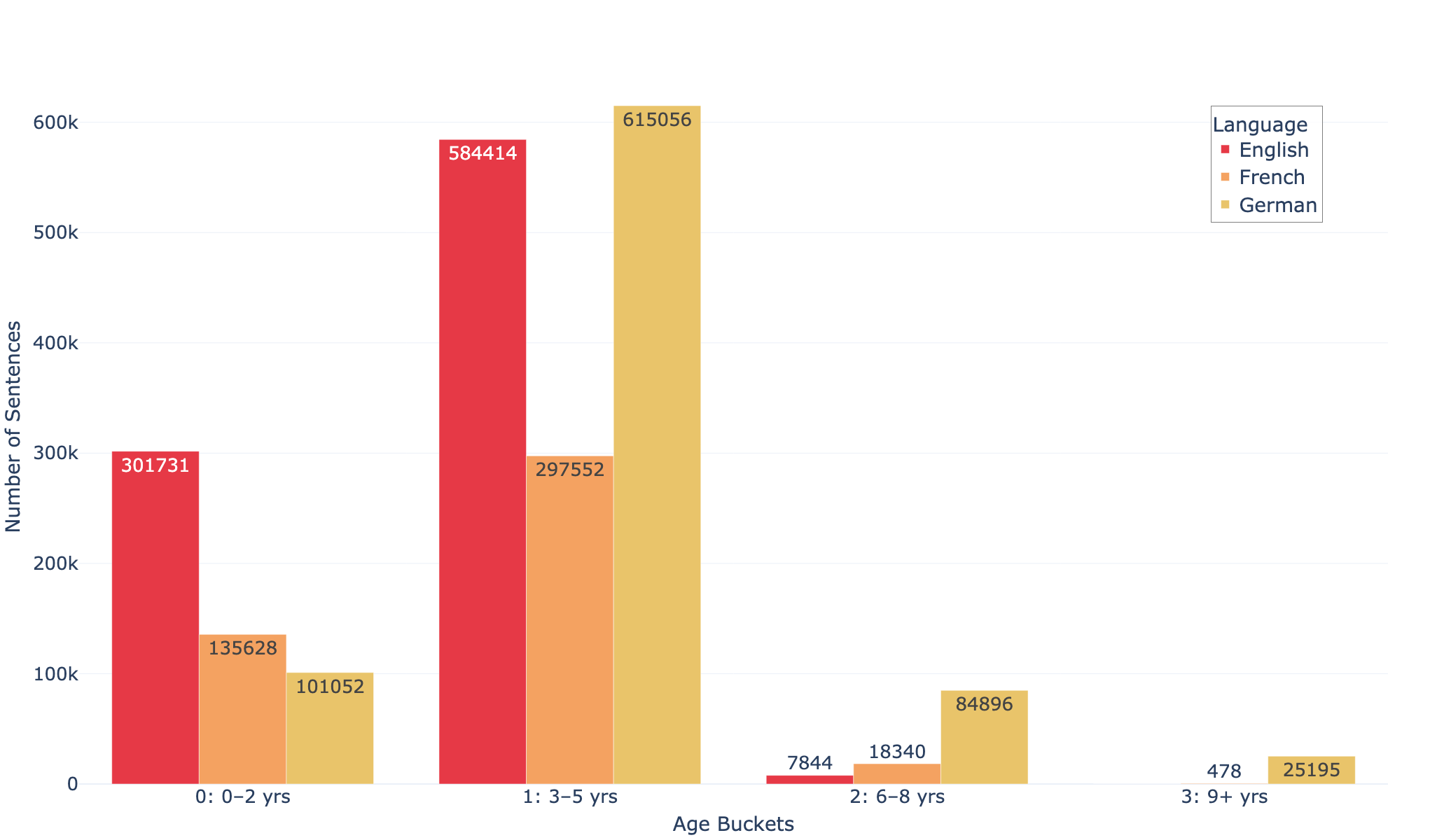}
    \caption{Sentence Count per Age Group in the three CHILDES datasets}
     \label{fig:sentence_length_count_age}
 \end{figure*}

As detailed in the main text, a subset of Wikipedia is selected for each language to closely match the token count of the corresponding CHILDES data. For English, we follow \citet{huebner-etal-2021-babyberta} and used \texttt{wikipedia1.txt} from their repository; for German, the corpus \texttt{gwlms/dewiki-20230701-nltk-corpus} was employed; and for French, we rely on \texttt{asi/wikitext\_fr}.
We report the differences between the two data types in the three target languages in terms of word frequency in~\Cref{fig:word_dist_lang} and sentence length in~\Cref{fig:sentence_length_dist_lang}. Additionally, as mentioned in the main text, since the age range covered by the CHILDES corpus varies across languages, in ~\Cref{fig:sentence_length_count_age} we display the total number of utterances directed at children of different ages for each CHILDES split. To generate the bins shown in Figure~\ref{fig:word_dist_lang}, we use the same strategy adopted for the new evaluation methodology described in~\Cref{fit_clams}, where the binning is done using uniformly spaced bins at a logarithmic scale, to account for the Zipfian nature of word frequencies. 

Moreover, ~\Cref{fig:interrogatives} provides a quantitative summary of the proportion of sentences classified as interrogatives in the two datasets. It clearly shows that interrogative sentences are substantially more frequent in CDL compared to Wikipedia.

\begin{table}[h!]
\centering
\begin{tabular}{|c|c|c|}
\hline
\textbf{Language} & \textbf{CHILDES} & \textbf{Wikipedia} \\ \hline
English & 39.84\% & 0.07\% \\ \hline
French & 31.28\% & 0.28\% \\ \hline
German & 28.93\% & 0.09\% \\ \hline
\end{tabular}
\caption{Comparison of interrogative clauses in CHILDES and Wikipedia datasets across languages.}

\label{fig:interrogatives}
\end{table}

\section{Models Details}
Table~\ref{tab:model-config} summarizes the configuration of the MLM and CLM models used in our experiments. We align the hyperparameters as closely as possible between the two architectures.

\begin{figure*}[h]
    \centering
    \text{ENGLISH}\\[0.3em]
    \begin{subfigure}[t]{0.23\textwidth}
        \centering
        \includegraphics[width=\linewidth]{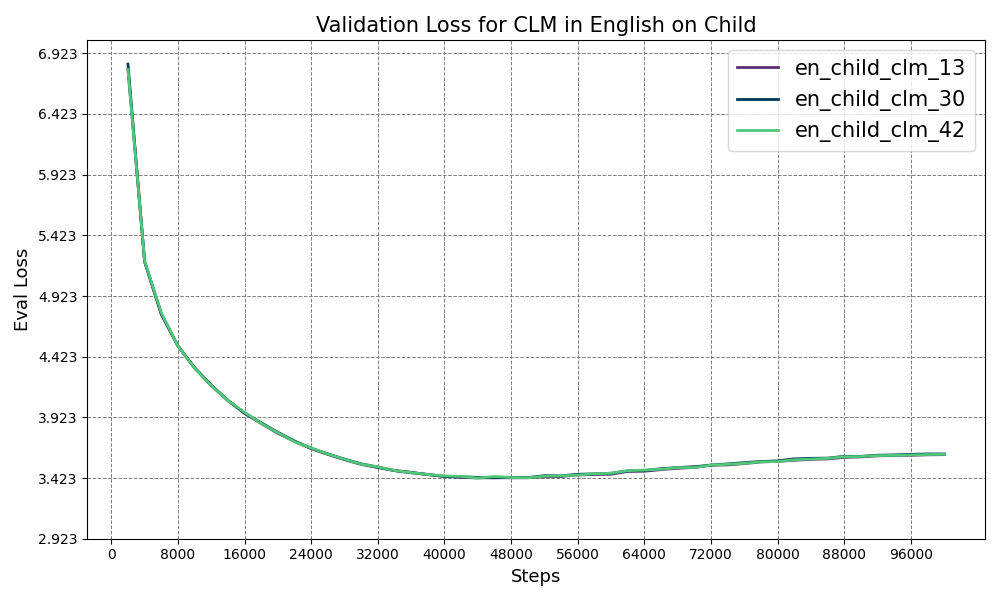}
    \end{subfigure}
    \hfill
    \begin{subfigure}[t]{0.23\textwidth}
        \centering
        \includegraphics[width=\linewidth]{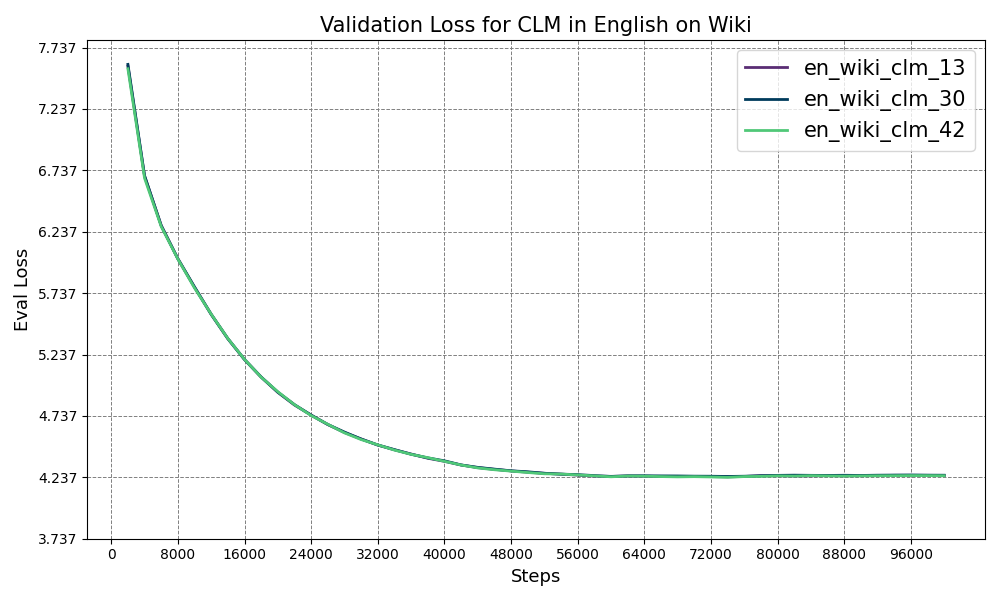}
    \end{subfigure}
    \hfill
    \begin{subfigure}[t]{0.23\textwidth}
        \centering
        \includegraphics[width=\linewidth]{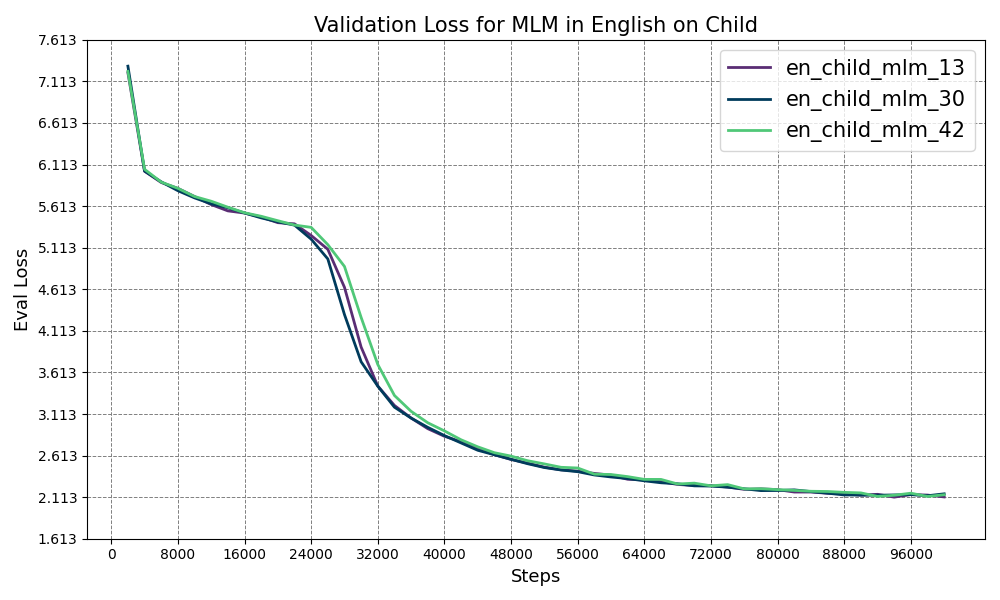}
    \end{subfigure}
    \hfill
    \begin{subfigure}[t]{0.23\textwidth}
        \centering
        \includegraphics[width=\linewidth]{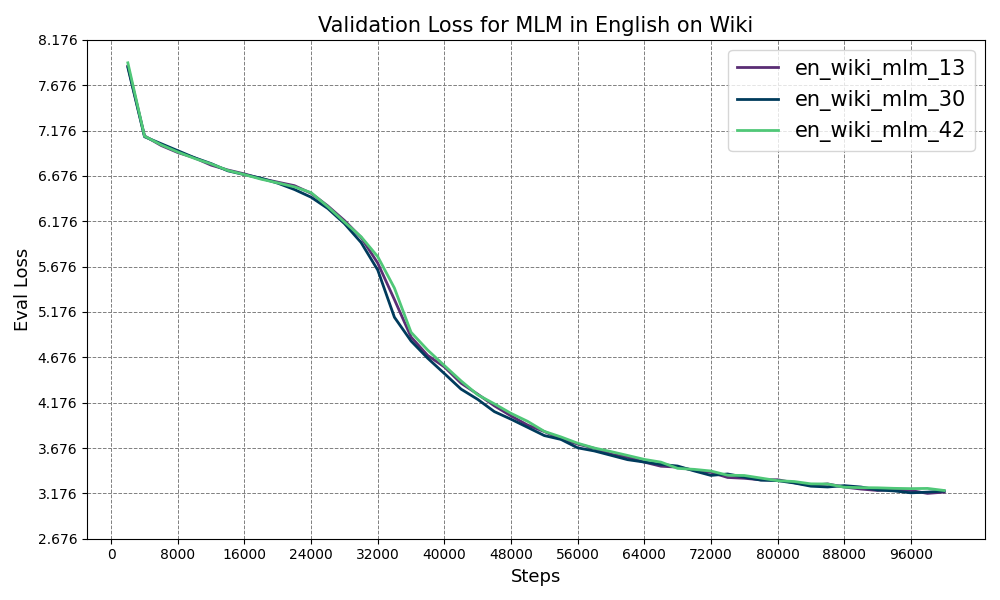}
    \end{subfigure}

    \vspace{2em}

    % FRENCH
    \text{FRENCH}\\[0.3em]
    \begin{subfigure}[t]{0.23\textwidth}
        \centering
        \includegraphics[width=\linewidth]{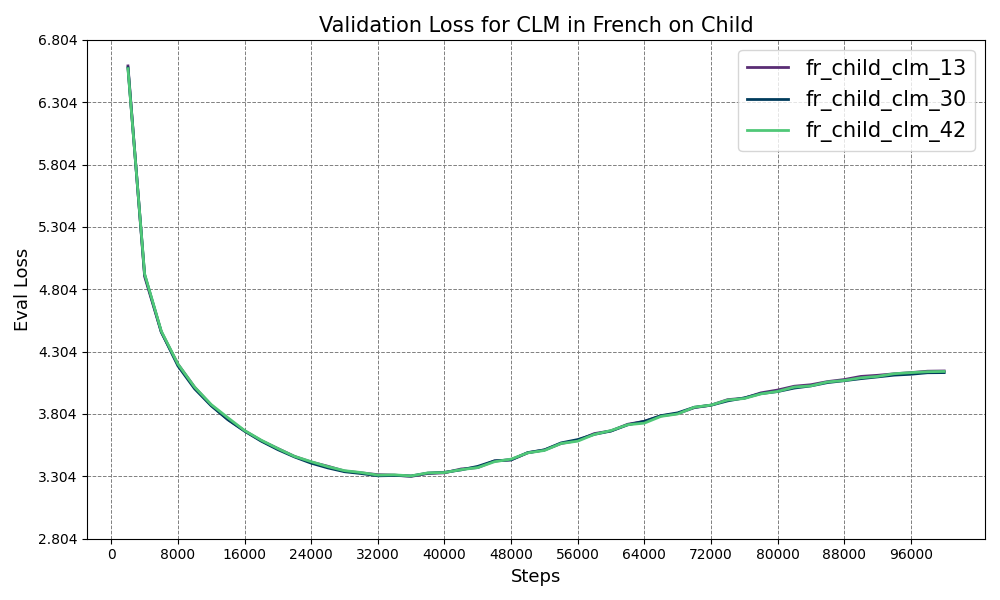}
    \end{subfigure}
    \hfill
    \begin{subfigure}[t]{0.23\textwidth}
        \centering
    
        \includegraphics[width=\linewidth]{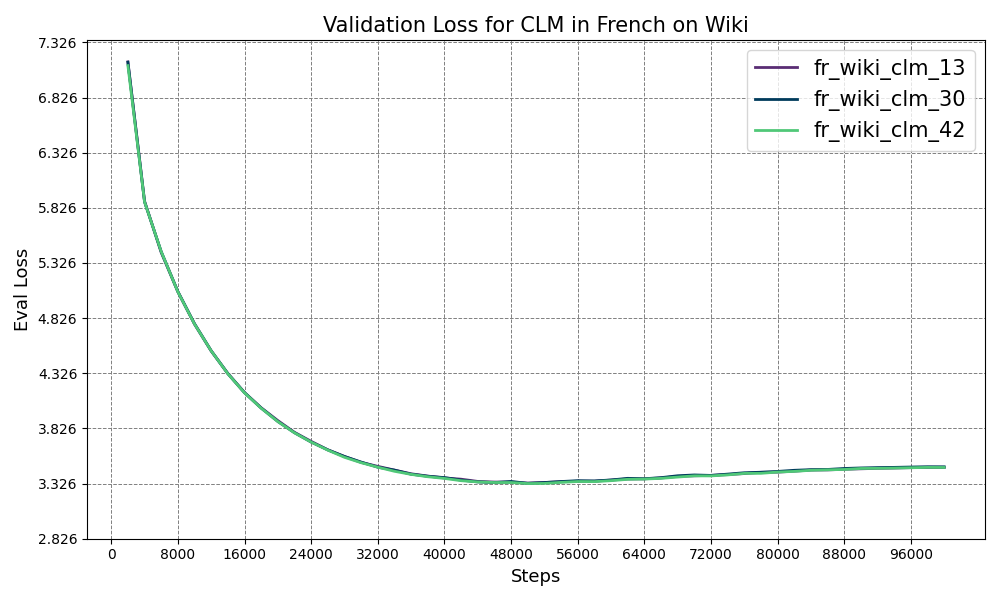}
    \end{subfigure}
    \hfill
    \begin{subfigure}[t]{0.23\textwidth}
        \centering
        \includegraphics[width=\linewidth]{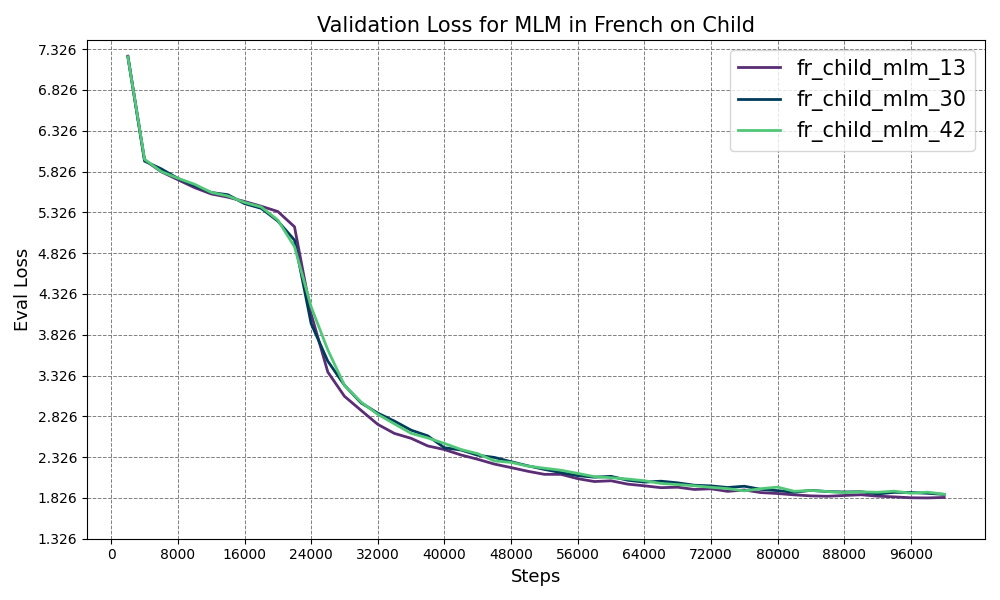}
    \end{subfigure}
    \hfill
    \begin{subfigure}[t]{0.23\textwidth}
        \centering
        \includegraphics[width=\linewidth]{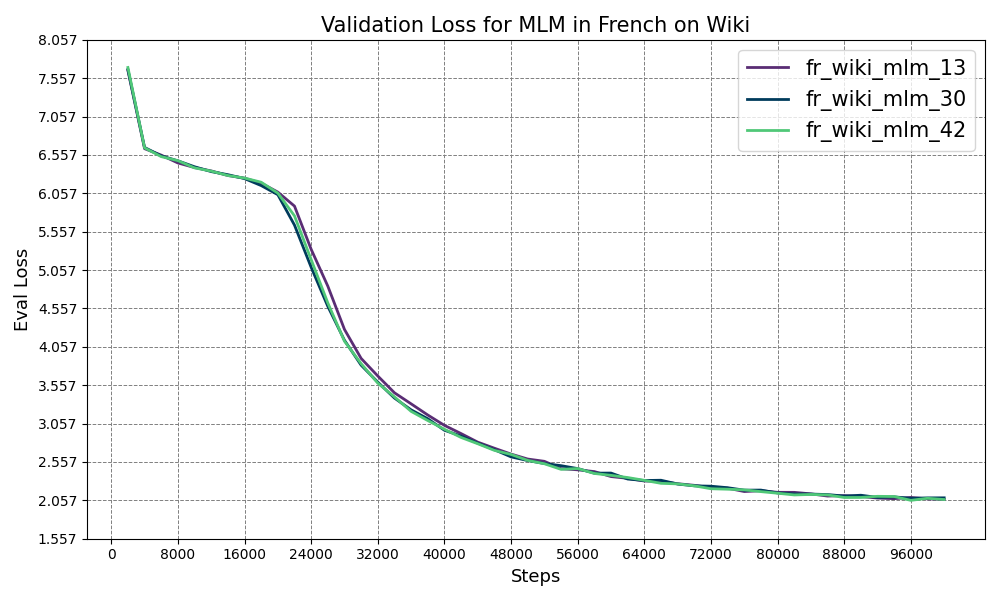}
    \end{subfigure}

    \vspace{2em}

    % GERMAN
    \text{GERMAN}\\[0.3em]
    \begin{subfigure}[t]{0.23\textwidth}
        \centering
        \includegraphics[width=\linewidth]{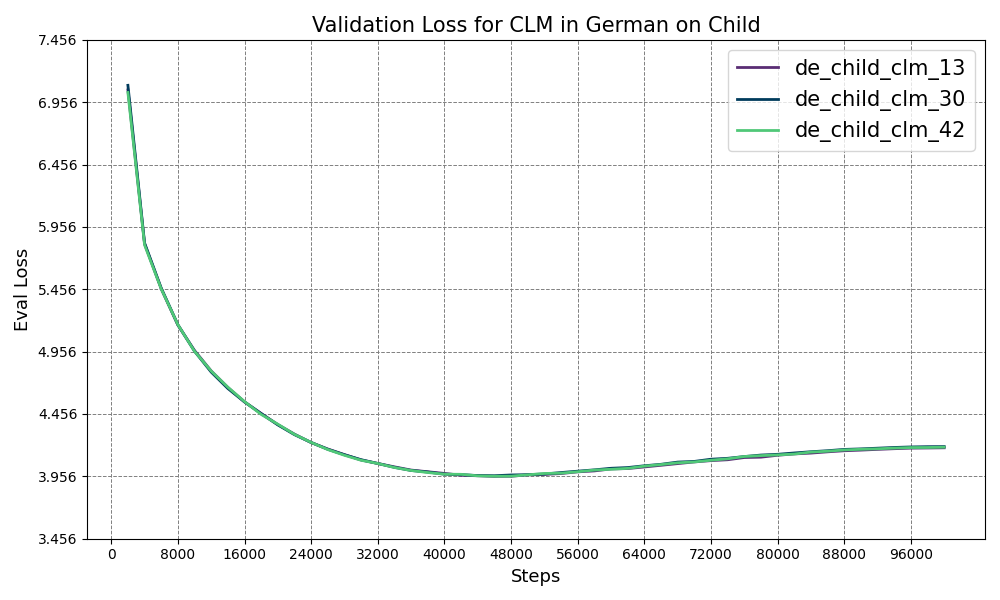}
    \end{subfigure}
    \hfill
    \begin{subfigure}[t]{0.23\textwidth}
        \centering
        \includegraphics[width=\linewidth]{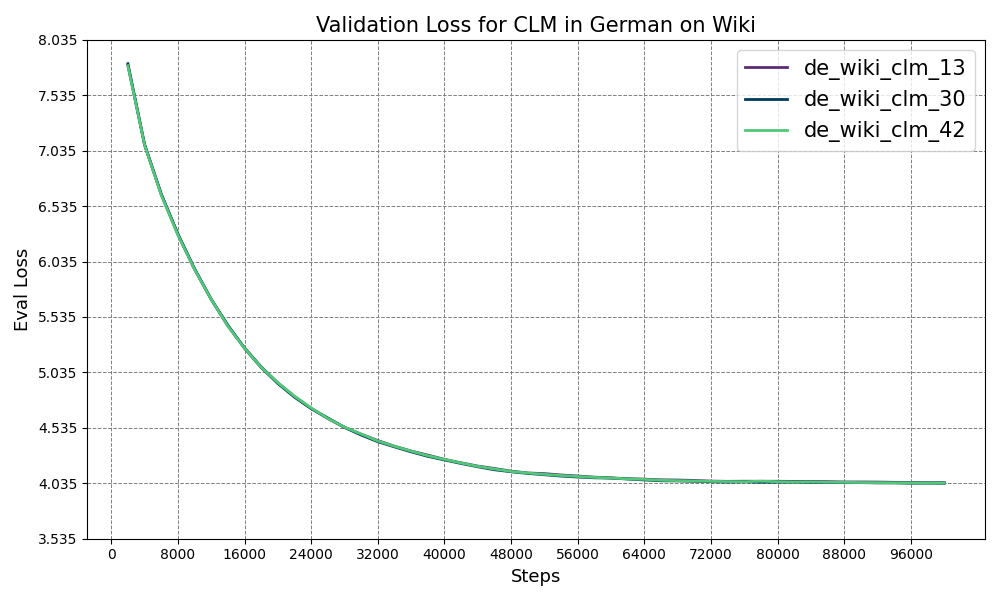}
    \end{subfigure}
    \hfill
    \begin{subfigure}[t]{0.23\textwidth}
        \centering
        \includegraphics[width=\linewidth]{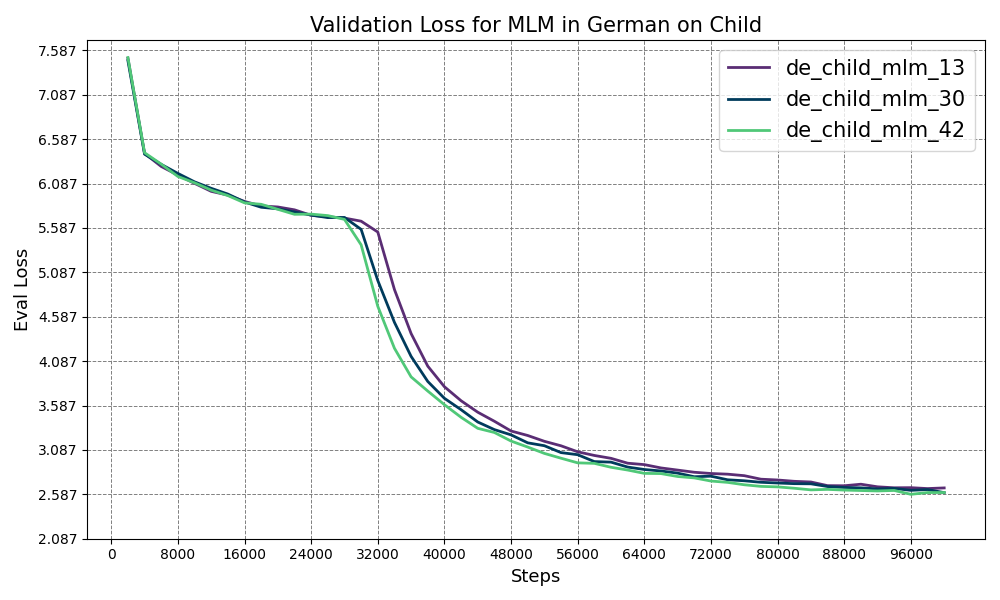}
    \end{subfigure}
    \hfill
    \begin{subfigure}[t]{0.23\textwidth}
        \centering
        \includegraphics[width=\linewidth]{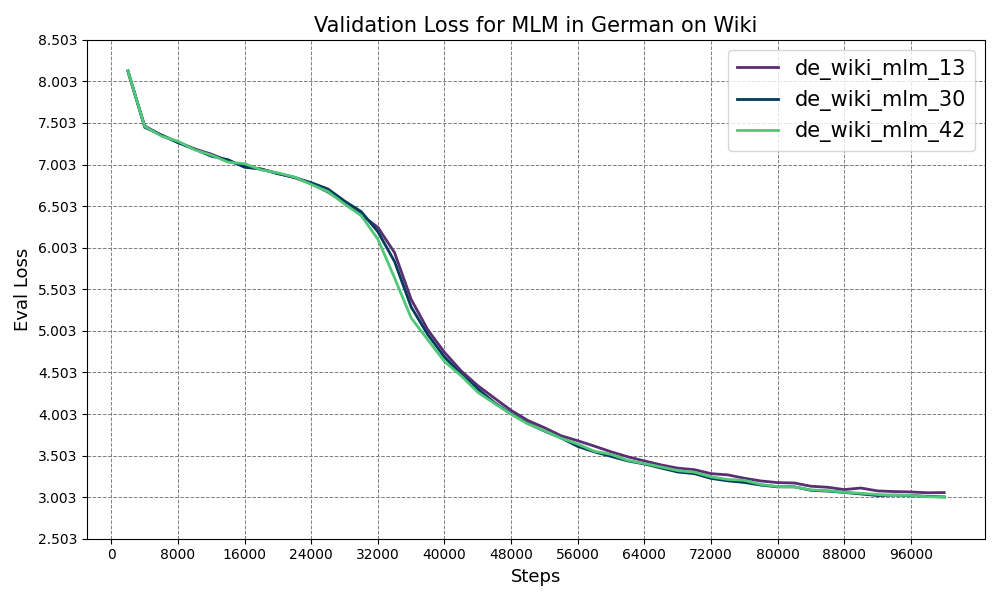}
    \end{subfigure}

    \caption{Validation perplexity curves for CLM and MLM models trained on CHILDES and Wikipedia corpora across English, German, and French.}
    \label{fig:validation_losses}
\end{figure*}

Figure~\ref{fig:validation_losses} presents the validation perplexity curves for MLM and CLM models trained on CHILDES and Wikipedia corpora across English, French and German. A clear pattern of earlier overfitting emerges for CLMs trained on CHILDES, with validation perplexity increasing after fewer training steps compared to their Wikipedia-trained counterparts. 

Table~\ref{tab:best-checkpoints} reports the CLM checkpoints selected for each language and dataset based on validation perplexity before overfitting, which we used for evaluation on both the existing benchmarks and our FIT-CLAMS minimal pairs.

\label{sec:models}
\begin{table}[ht]
\centering
\renewcommand{\arraystretch}{1.3}
\small
\begin{tabular}{lll}
\toprule
\textbf{Hyperparameter} & \textbf{MLM} & \textbf{CLM} \\
\midrule
Architecture            & RoBERTa        & GPT-2     \\
\hline
Layers                  & 8            & 8           \\
\hline
Attention heads         & 8            & 8           \\
\hline
Intermediate size       & 2048         & 2048        \\
\hline
Max seq. length         & 512          & 512         \\
\hline
Objective               & Masked LM    & Causal LM   \\
\hline
Total parameters        & 12.7M    & 14.8M   \\
\hline
Learning Rate           & 0.0001    & 0.0001   \\
\hline
lr\_scheduler\_type  & linear    & linear   \\
\hline
Training Batch Size   & 16    & 16   \\
\hline
Evaluation Batch Size   & 16    & 16   \\
\hline
Gradient Accumulation Step  & 2    & 2   \\
\hline
\bottomrule
\end{tabular}
\caption{Comparison of CLM and MLM model configurations.}
\vspace{-1em}
\label{tab:model-config}
\end{table}

\begin{table}[ht]
\centering
\small
\renewcommand{\arraystretch}{1.2}
\begin{tabular}{l|l|l}
\textbf{Language} & \textbf{CHILDES} & \textbf{Wikipedia} \\
\hline
EN & ckpt-48000 & ckpt-64000 \\
\hline
FR  & ckpt-36000 & ckpt-44000 \\
\hline
DE  & ckpt-48000 & ckpt-64000 \\
\hline
\end{tabular}
\caption{CLM checkpoints per language and training dataset that we used for evaluation on existing benchmarks and on FIT-CLAMS.}
\label{tab:best-checkpoints}
\end{table}

\section{Accuracy Results of CLMs and MLMs on existing benchmarks}
\label{sec:eval_bench_exist}
\begin{figure*}[h]
     \centering
     \includegraphics[width=0.7\linewidth]{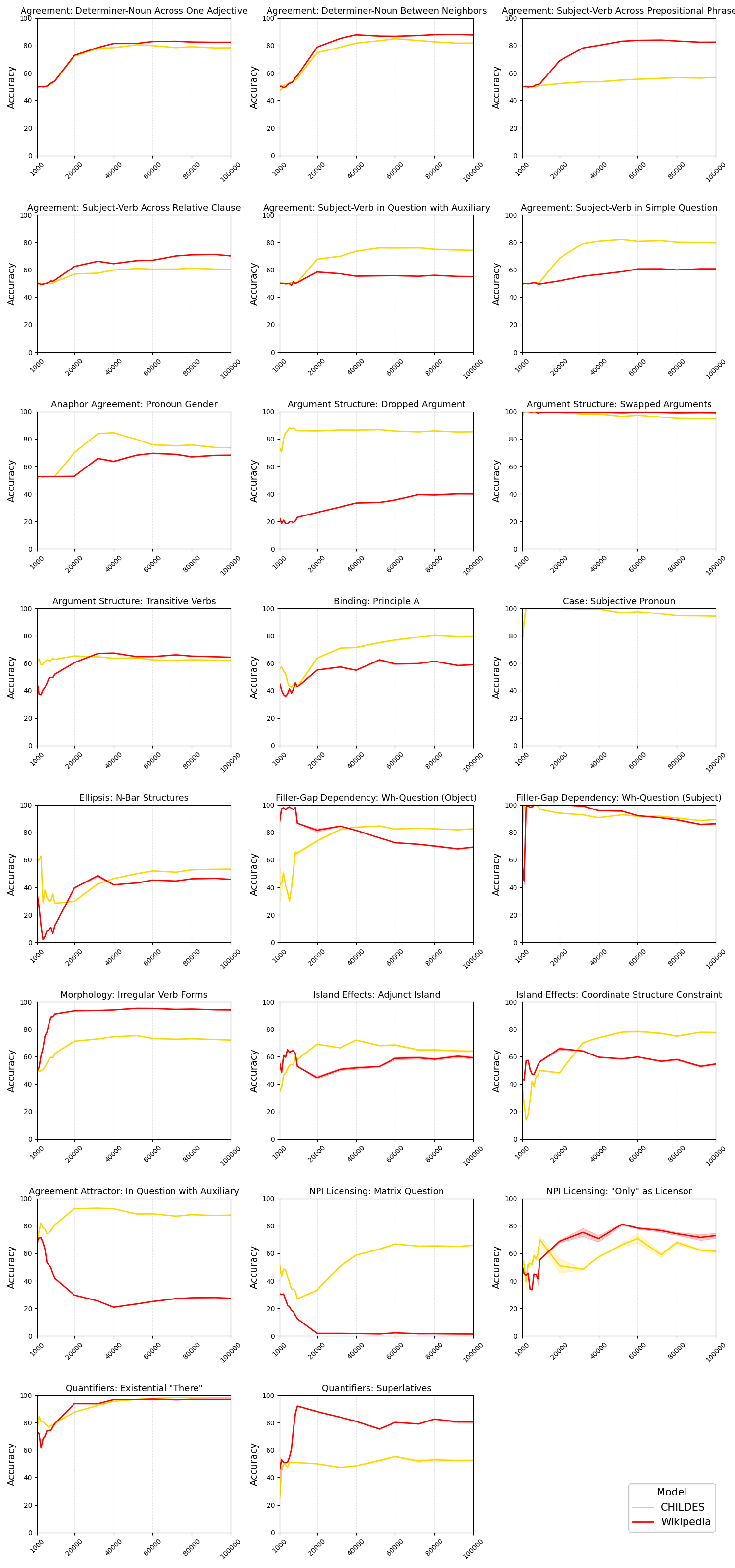}
    \caption{CLM models' accuracy curves on Zorro.}
     \label{fig:curves_zorro}
 \end{figure*}

\begin{figure*}[h]
     \centering
     \includegraphics[width=0.7\linewidth]{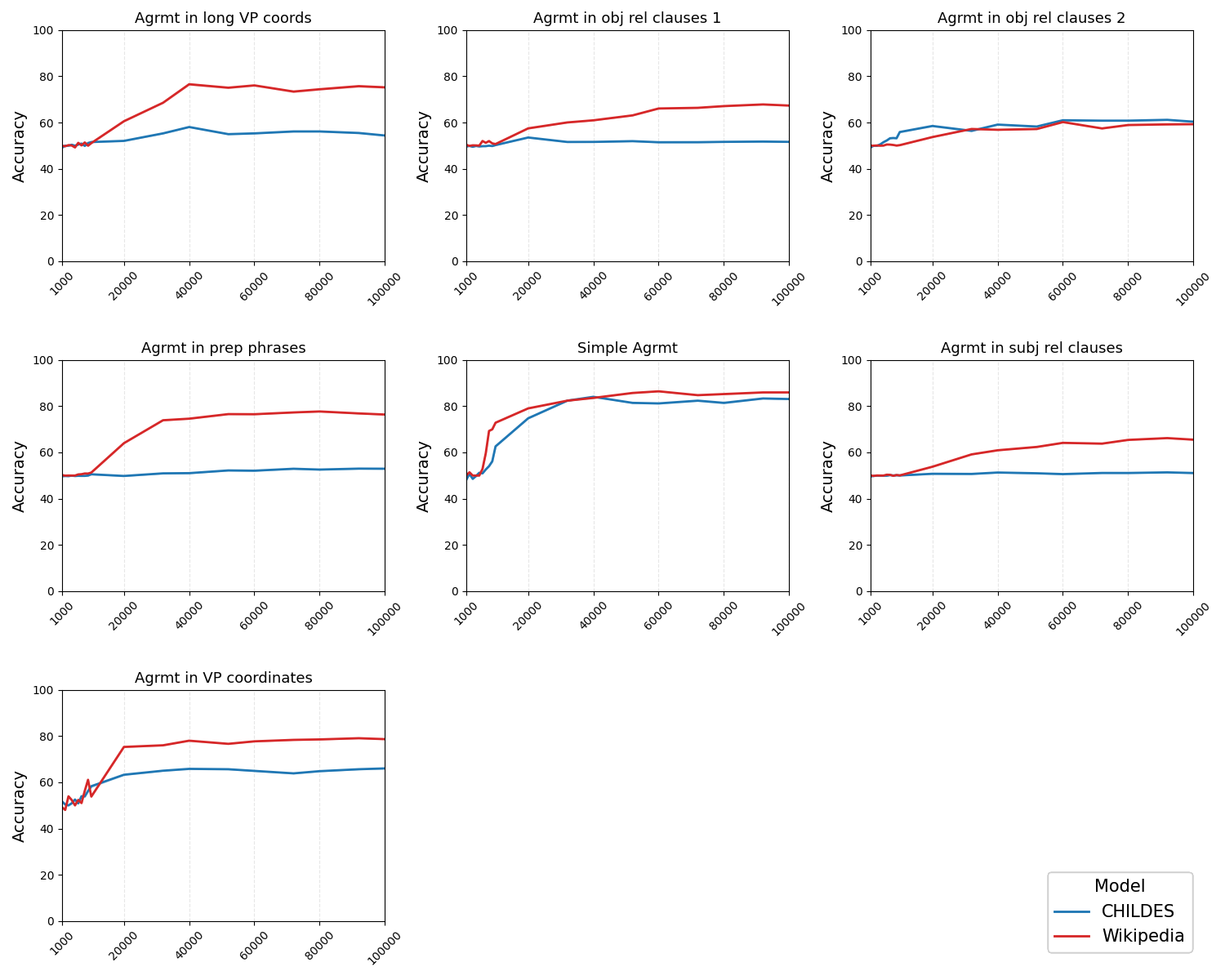}
    \caption{English CLM models' accuracy curves on CLAMS.}
     \label{fig:english_clams_lc}
 \end{figure*}

\begin{figure*}[h]
     \centering
     \includegraphics[width=0.7\linewidth]{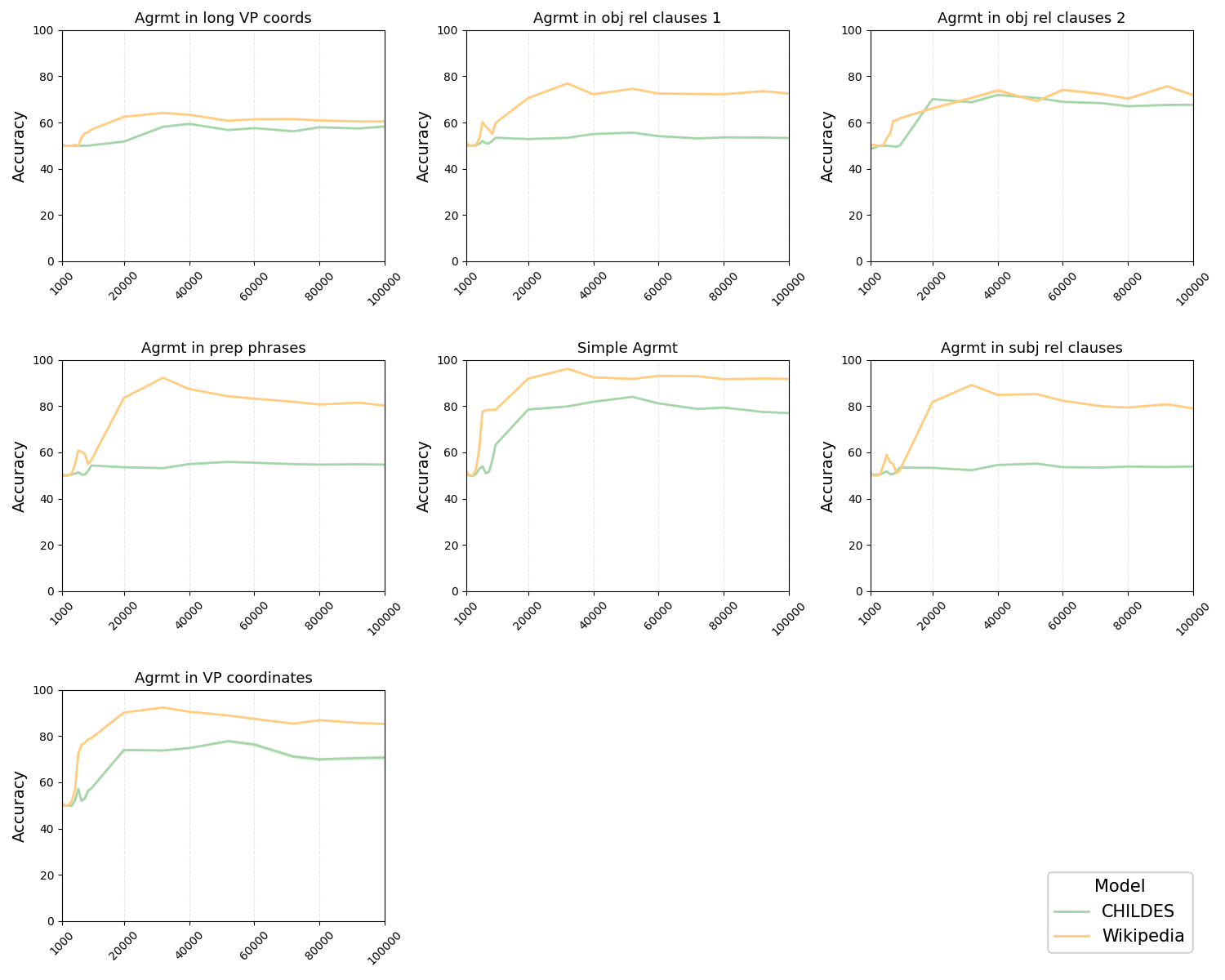}
    \caption{French CLM models' accuracy curves on CLAMS.}
     \label{fig:french_clams_lc}
 \end{figure*}

\begin{figure*}[h]
     \centering
     \includegraphics[width=0.7\linewidth]{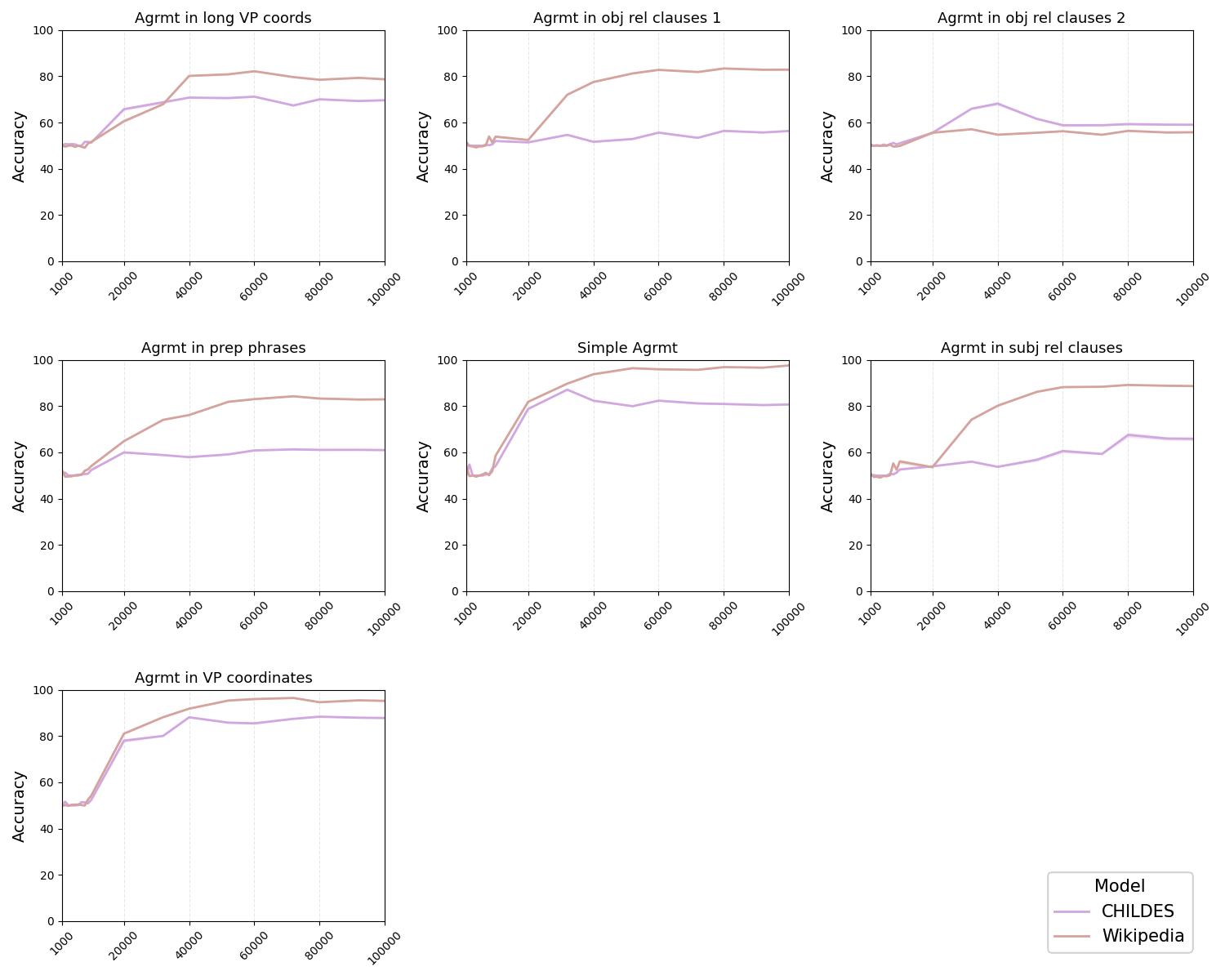}
    \caption{German CLM models' accuracy curves on CLAMS.}
     \label{fig:german_clams_lc}
 \end{figure*}

Models are evaluated on each benchmark across 19 training checkpoints: 10 selected from the first 10\% of training steps and 9 from the remaining 90\%. This selection strategy allows for a more detailed examination of the early-stage learning trajectories. 
~\Cref{fig:curves_zorro} refers to Zorro accuracy learning curves across steps for our CHILDES- and Wikipedia-models trained on English. Instead,  ~\Cref{fig:english_clams_lc}-~\ref{fig:french_clams_lc}-~\ref{fig:german_clams_lc} show the accuracy learning curves on CLAMS of our CLMs trained on the three language of interest.

Tables~\ref{tab:clm_results_zorro}-~\ref{tab:mlm_results_zorro} report the accuracies of the two models types (CLM and MLM) trained on the two different datasets (CHILDES vs Wikipedia) for each paradigm targeted in Zorro. Paradigms involving questions are highlighted in bold and with color. For CLMs, 8 out of 13 paradigms where CHILDES outperforms Wikipedia include questions, whereas the advantage for question-related paradigms is less pronounced in the case of the MLMs (only 6 out of 12).

\begin{table*}[ht]
\centering
\tiny
\renewcommand{\arraystretch}{1.3}
\begin{minipage}{0.48\textwidth}
\centering
\begin{tabular}{lcc}
\rowcolor{cherryred!20}
\textbf{Paradigm} & \textbf{CHILDES} & \textbf{Wiki} \\
\toprule
agr\_det\_noun\_across\_1\_adj & 0.813 & \textbf{0.816} \\
agr\_det\_noun\_between\_neighbors & 0.846 & \textbf{0.893} \\
\rowcolor{cherryred!30} \textbf{agreement\_subj\_verb\_in\_q\_with\_aux} & \textbf{0.743} & 0.574 \\
agr\_subj\_verb\_across\_prep\_phr & 0.546 & \textbf{0.831} \\
agr\_subj\_verb\_across\_relclause & 0.606 & \textbf{0.717} \\
\rowcolor{cherryred!30} \textbf{agr\_subj\_verb\_in\_simple\_q} & \textbf{0.799} & 0.581 \\
\rowcolor{cherryred!30} \textbf{anaphor\_agreement\_pronoun\_gender} & \textbf{0.823} & 0.678 \\
arg\_structure\_dropped\_arg & \textbf{0.861} & 0.429 \\
arg\_structure\_swapped\_args & 0.973 & \textbf{0.994} \\
\rowcolor{cherryred!30} \textbf{arg\_structure\_transitive} & \textbf{0.626} & 0.622 \\
binding\_principle\_a & \textbf{0.771} & 0.622 \\
case\_subj\_pronoun & 0.983 & \textbf{1.000} \\
ellipsis\_n\_bar & \textbf{0.475} & 0.466 \\
filler\_gap\_wh\_q\_object & \textbf{0.834} & 0.799 \\
filler\_gap\_wh\_q\_subject & 0.916 & \textbf{0.932} \\
irregular\_verb & 0.735 & \textbf{0.928} \\
\rowcolor{cherryred!30} \textbf{island\_effects\_adjunct\_island} & \textbf{0.665} & 0.546 \\
\rowcolor{cherryred!30} \textbf{island\_effects\_coord\_constraint} & \textbf{0.765} & 0.646 \\
\rowcolor{cherryred!30} \textbf{local\_attractor\_in\_q\_aux} & \textbf{0.912} & 0.314 \\
\rowcolor{cherryred!30} \textbf{npi\_licensing\_matrix\_question} & \textbf{0.648} & 0.078 \\
npi\_licensing\_only\_npi\_lic & 0.719 & \textbf{0.773} \\
quantifiers\_existential\_there & \textbf{0.957} & 0.956 \\
quantifiers\_superlative & 0.496 & \textbf{0.703} \\
\bottomrule
\end{tabular}
\caption{CLM scores on Zorro subparadigms. Question-related paradigms are emphasized with boldface and deeper highlighting.}
\label{tab:clm_results_zorro}
\end{minipage}
\hfill % Adds horizontal space between the minipages
\begin{minipage}{0.48\textwidth}
\centering
\begin{tabular}{lcc}
\rowcolor{orange!30}
\textbf{Paradigm} & \textbf{CHILDES} & \textbf{Wiki} \\
\toprule
agr\_det\_noun\_across\_1\_adj & 0.664 & \textbf{0.815} \\
agr\_det\_noun\_between\_neighbors & 0.726 & \textbf{0.899} \\
\rowcolor{orange!30} \textbf{agreement\_subj\_verb\_in\_q\_with\_aux} & \textbf{0.603} & 0.597 \\
agr\_subj\_verb\_across\_prep\_phr & 0.548 & \textbf{0.783} \\
agr\_subj\_verb\_across\_relclause & 0.559 & \textbf{0.641} \\
\rowcolor{orange!30} \textbf{agr\_subj\_verb\_in\_simple\_q} & \textbf{0.654} & 0.544 \\
\rowcolor{orange!30} \textbf{anaphor\_agreement\_pronoun\_gender} & \textbf{0.864} & 0.554 \\
arg\_structure\_dropped\_arg & \textbf{0.550} & 0.349 \\
arg\_structure\_swapped\_args & \textbf{0.705} & 0.555 \\
\rowcolor{orange!30} \textbf{arg\_structure\_transitive} & 0.527 & \textbf{0.555} \\
binding\_principle\_a & 0.805 & \textbf{0.856} \\
case\_subj\_pronoun & \textbf{0.862} & 0.848 \\
ellipsis\_n\_bar & \textbf{0.671} & 0.523 \\
filler\_gap\_wh\_q\_object & \textbf{0.852} & 0.797 \\
filler\_gap\_wh\_q\_subject & 0.828 & \textbf{0.971} \\
irregular\_verb & 0.634 & \textbf{0.921} \\
\rowcolor{orange!30} \textbf{island\_effects\_adjunct\_island} & \textbf{0.612} & 0.600 \\
\rowcolor{orange!30} \textbf{island\_effects\_coord\_constraint} & 0.572 & \textbf{0.883} \\
\rowcolor{orange!30} \textbf{local\_attractor\_in\_q\_aux} & \textbf{0.727} & 0.358 \\
\rowcolor{orange!30} \textbf{npi\_licensing\_matrix\_question} & \textbf{0.260} & 0.022 \\
npi\_licensing\_only\_npi\_lic & 0.707 & \textbf{0.791} \\
quantifiers\_existential\_there & \textbf{0.970} & 0.922 \\
quantifiers\_superlative & 0.376 & \textbf{0.670} \\
\bottomrule
\end{tabular}
\caption{MLM scores on Zorro subparadigms. Question-related paradigms are emphasized with boldface and deeper highlighting.}
\label{tab:mlm_results_zorro}
\end{minipage}
\end{table*}

\begin{figure*}[htbp]
    \centering
    \small

    \begin{subfigure}[t]{0.95\textwidth}
        \centering
        \includegraphics[width=\linewidth]{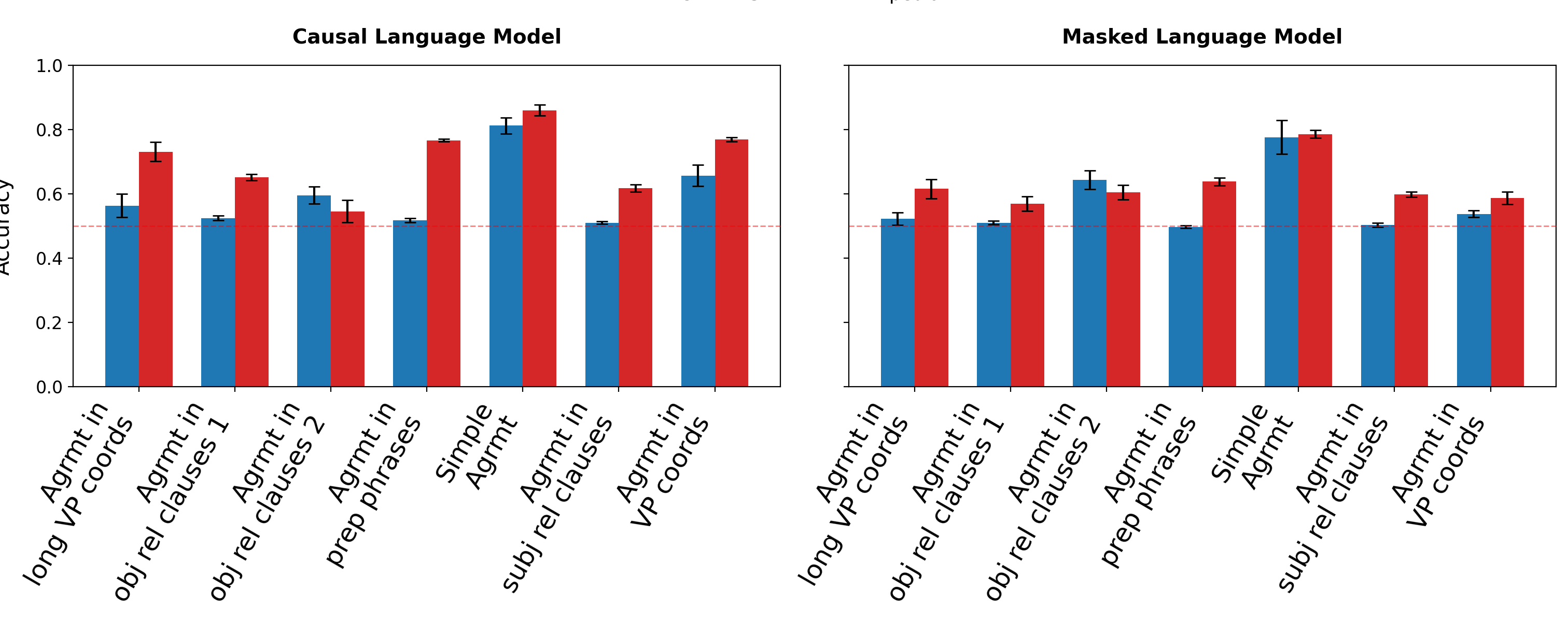}
        \label{fig:en_dual}
    \end{subfigure}
    \vspace{1em}

    \begin{subfigure}[t]{0.95\textwidth}
        \centering
        \includegraphics[width=\linewidth]{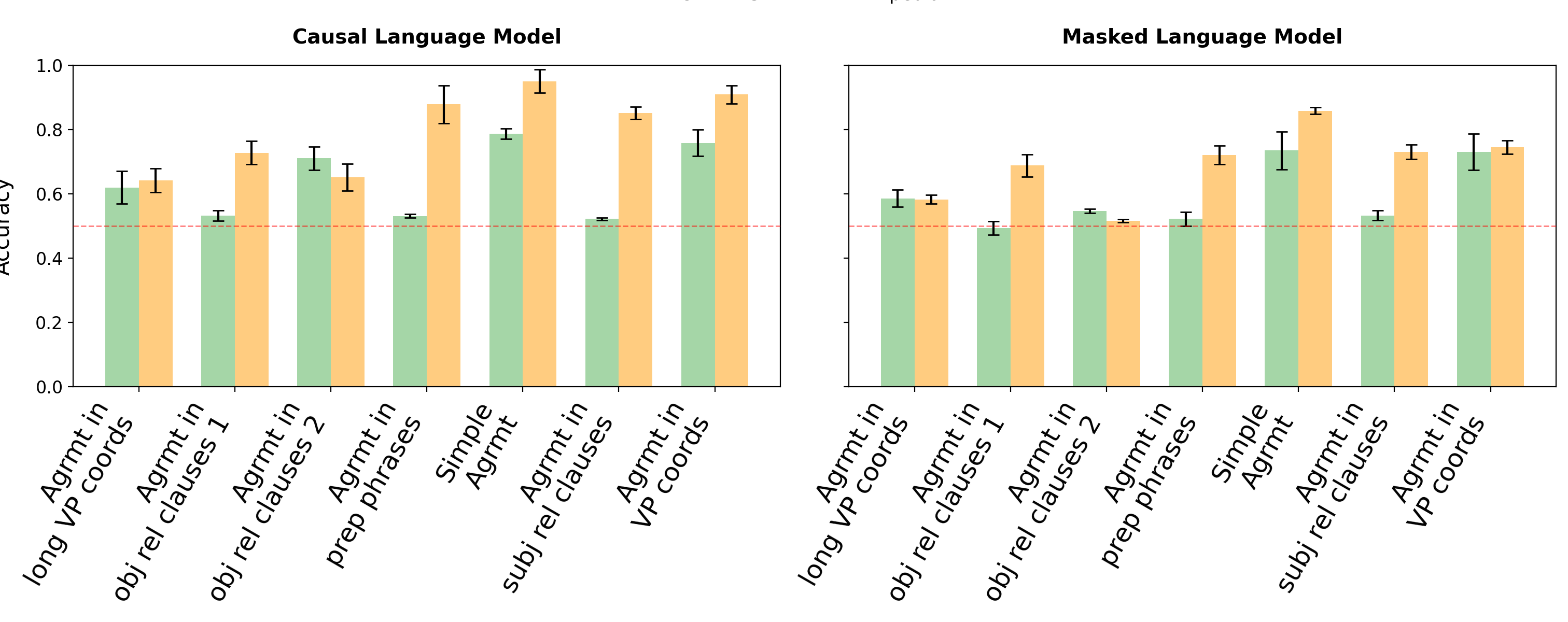}
        \label{fig:fr_dual}
    \end{subfigure}
    \vspace{1em}

    \begin{subfigure}[t]{0.95\textwidth}
        \centering
        \includegraphics[width=\linewidth]{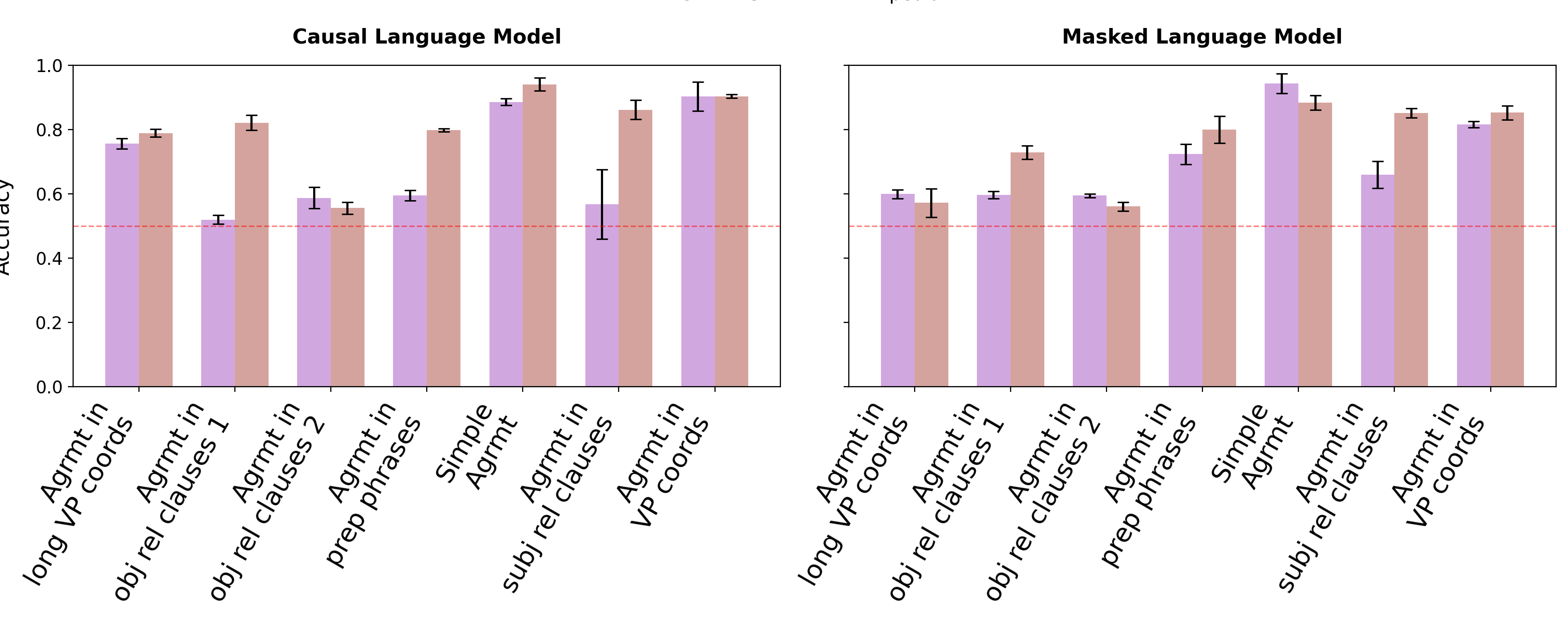}
        \label{fig:de_dual}
    \end{subfigure}

    \caption{Accuracy scores per paradigm for CLM and MLM across languages on CLAMS.}
    \label{fig:clm_mlm_comparison}
\end{figure*}

Figure~\ref{fig:clm_mlm_comparison} illustrates the performance of our two model architectures—CLM and MLM—when evaluated on the CLAMS benchmark, providing a visual summary of their accuracy across languages and paradigms. The highest accuracy is observed in the simple agreement paradigm— the least complex, involving only an article, a noun and a verb. As agreement complexity increases, overall performance declines, yet the Wikipedia-trained models continue to hold an advantage, except in the agreement within relative clauses (Agrmt in obj rel clauses 2). 

\subsection{Evaluating Previous Models in the Literature}
\label{sec:test_huebner_models}

To validate our evaluation pipeline, we applied it to models released by \citet{huebner-etal-2021-babyberta}, allowing for a direct comparison with existing findings in the literature. Specifically, we evaluated two versions of their model trained on AO-CHILDES: one that uses no unmasking probability, and the other that uses the standard unmasking probability value typically employed in MLM training. In both cases, the average accuracies across all paradigms on the Zorro benchmark diverge from the results reported in their original paper. The first model (unmasking probability = 0) achieves an average accuracy of 66\%, while the second (standard unmasking probability) scores 68\%.

\section{FIT-CLAMS Minimal Pairs Generation}
\label{sec:new_clams}

\begin{figure*}[htbp]
    \centering

    % Row 1
    \begin{subfigure}[t]{0.22\textwidth}
        \centering
        \includegraphics[width=0.9\linewidth]{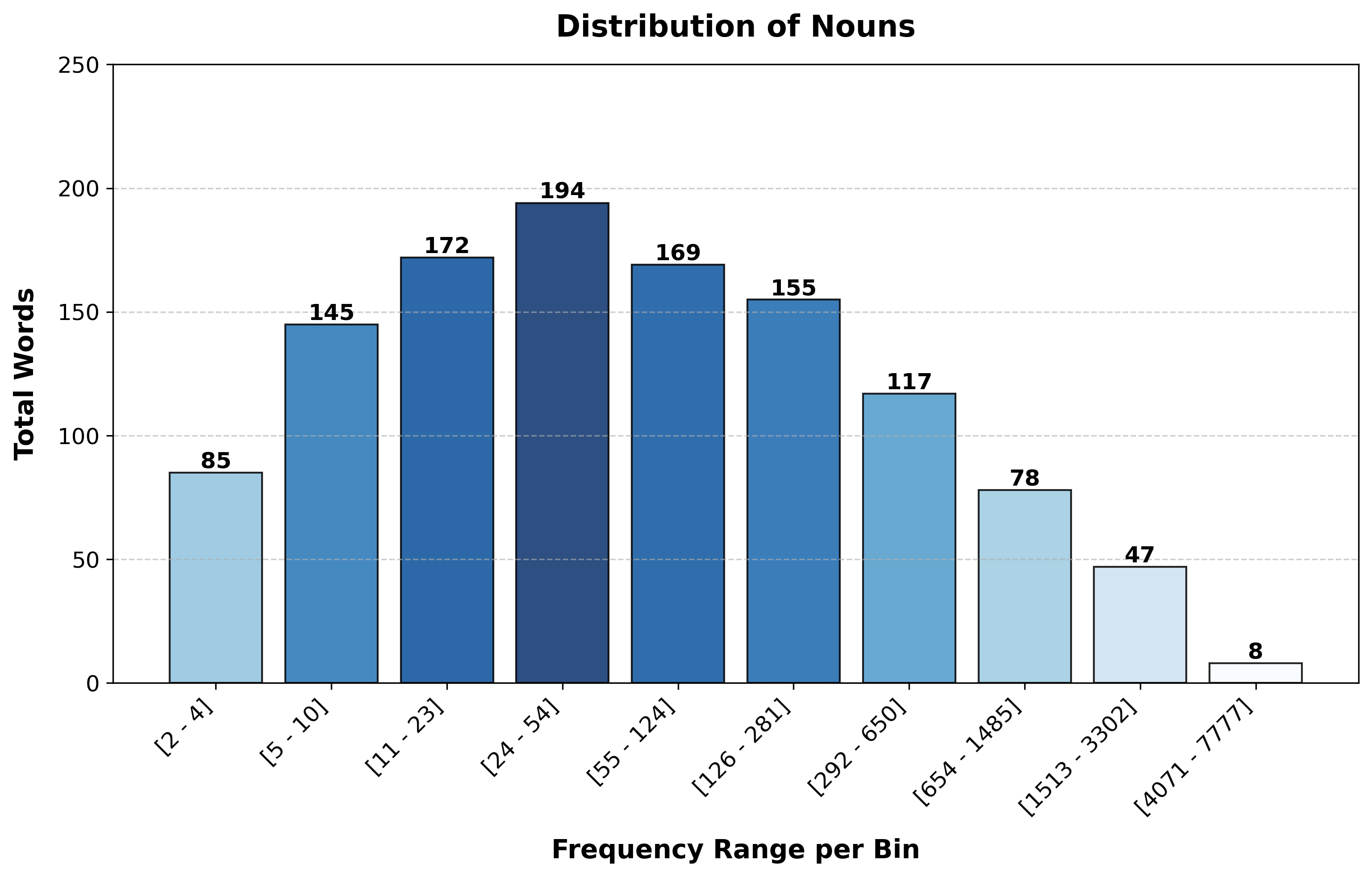}
        \caption{ENG - CHILDES}
    \end{subfigure}
    \hfill
    \begin{subfigure}[t]{0.22\textwidth}
        \centering
        \includegraphics[width=0.9\linewidth]{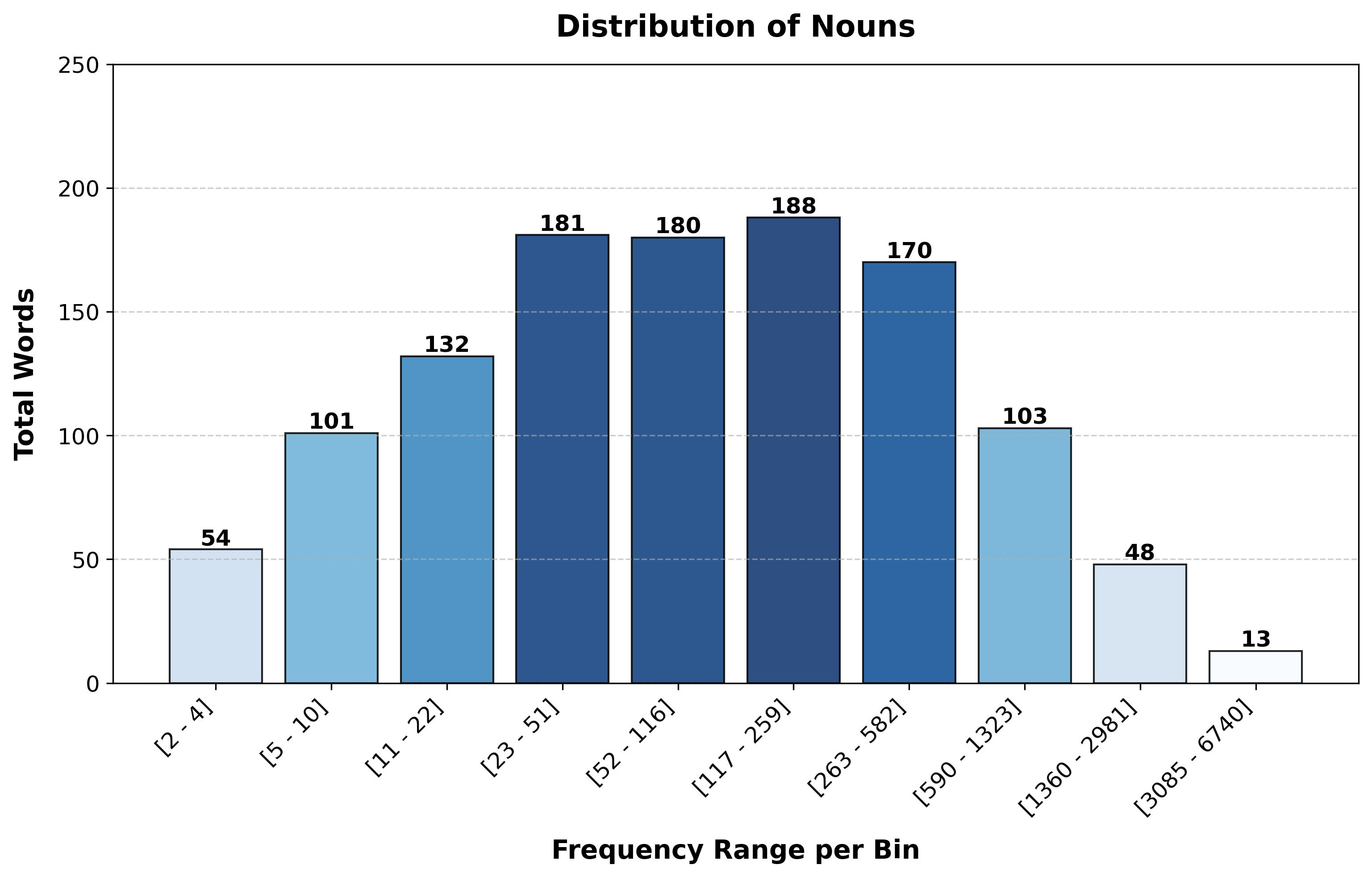}
        \caption{ENG - Wiki}
    \end{subfigure}
    \hfill
    \begin{subfigure}[t]{0.22\textwidth}
        \centering
        \includegraphics[width=0.9\linewidth]{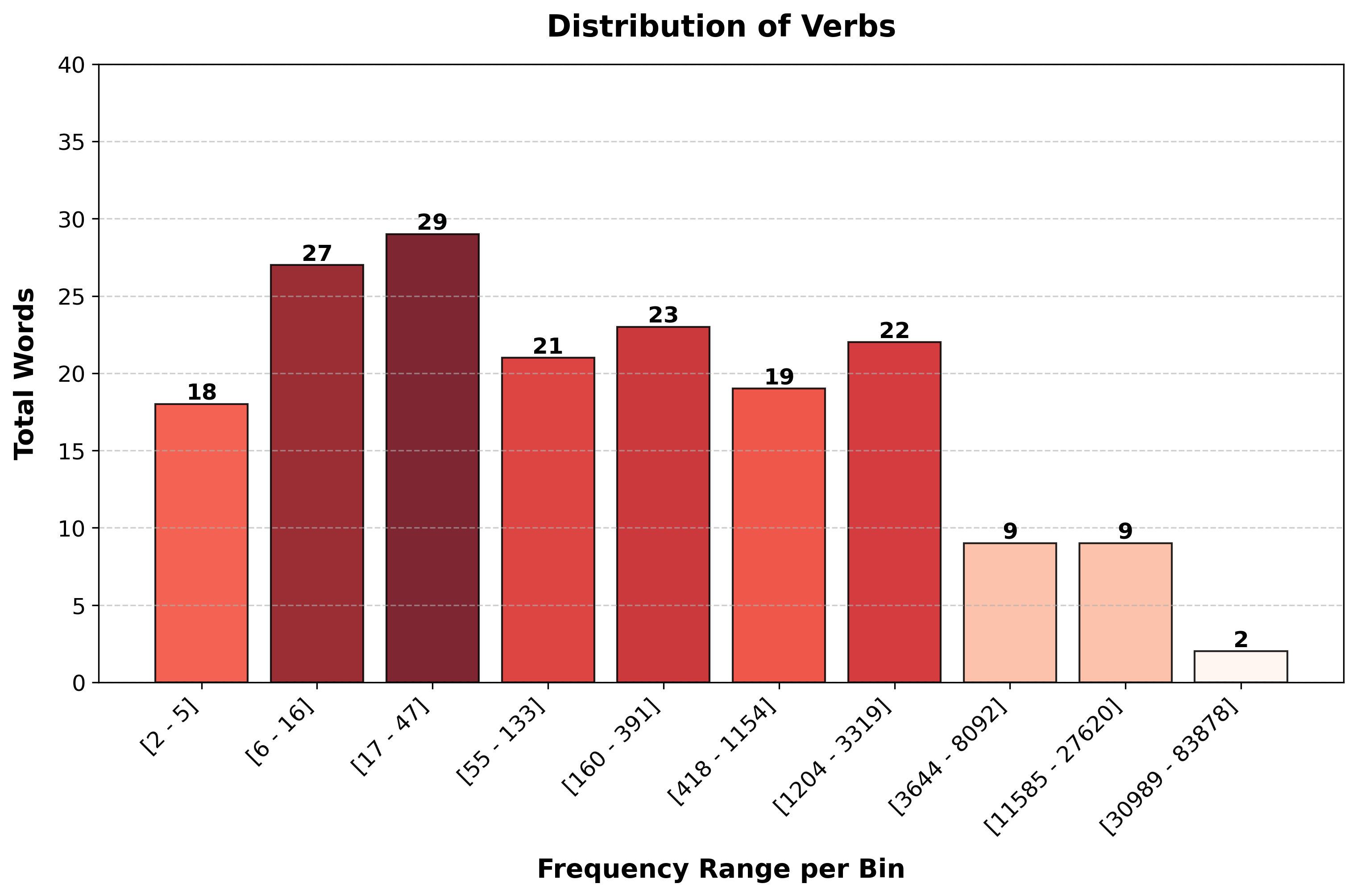}
        \caption{ENG - CHILDES}
    \end{subfigure}
    \hfill
    \begin{subfigure}[t]{0.22\textwidth}
        \centering
        \includegraphics[width=0.9\linewidth]{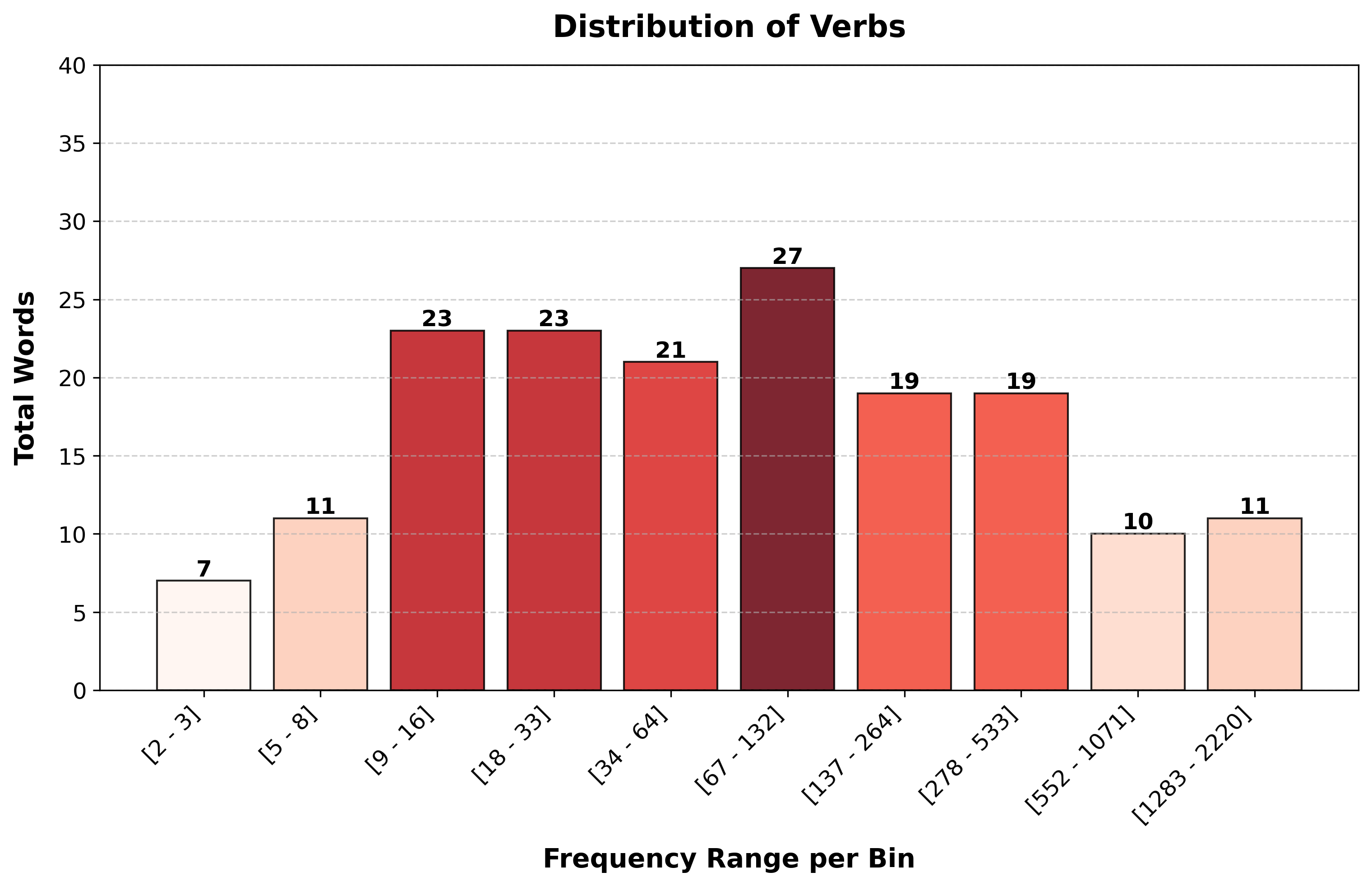}
        \caption{ENG - Wiki}
    \end{subfigure}

    \vspace{1em}

    % Row 2
    \begin{subfigure}[t]{0.22\textwidth}
        \centering
        \includegraphics[width=0.9\linewidth]{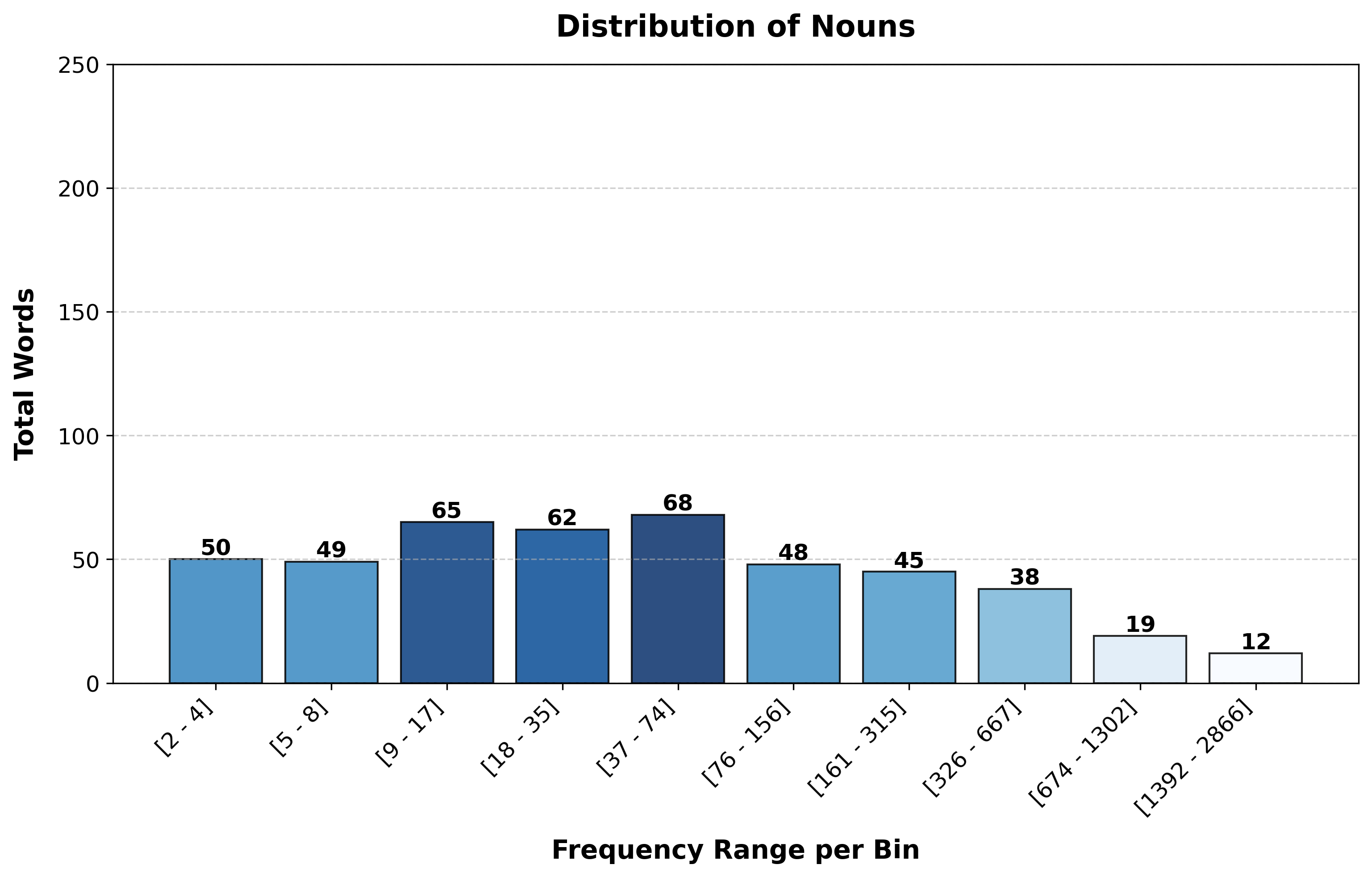}
        \caption{FR - CHILDES}
    \end{subfigure}
    \hfill
    \begin{subfigure}[t]{0.22\textwidth}
        \centering
        \includegraphics[width=0.9\linewidth]{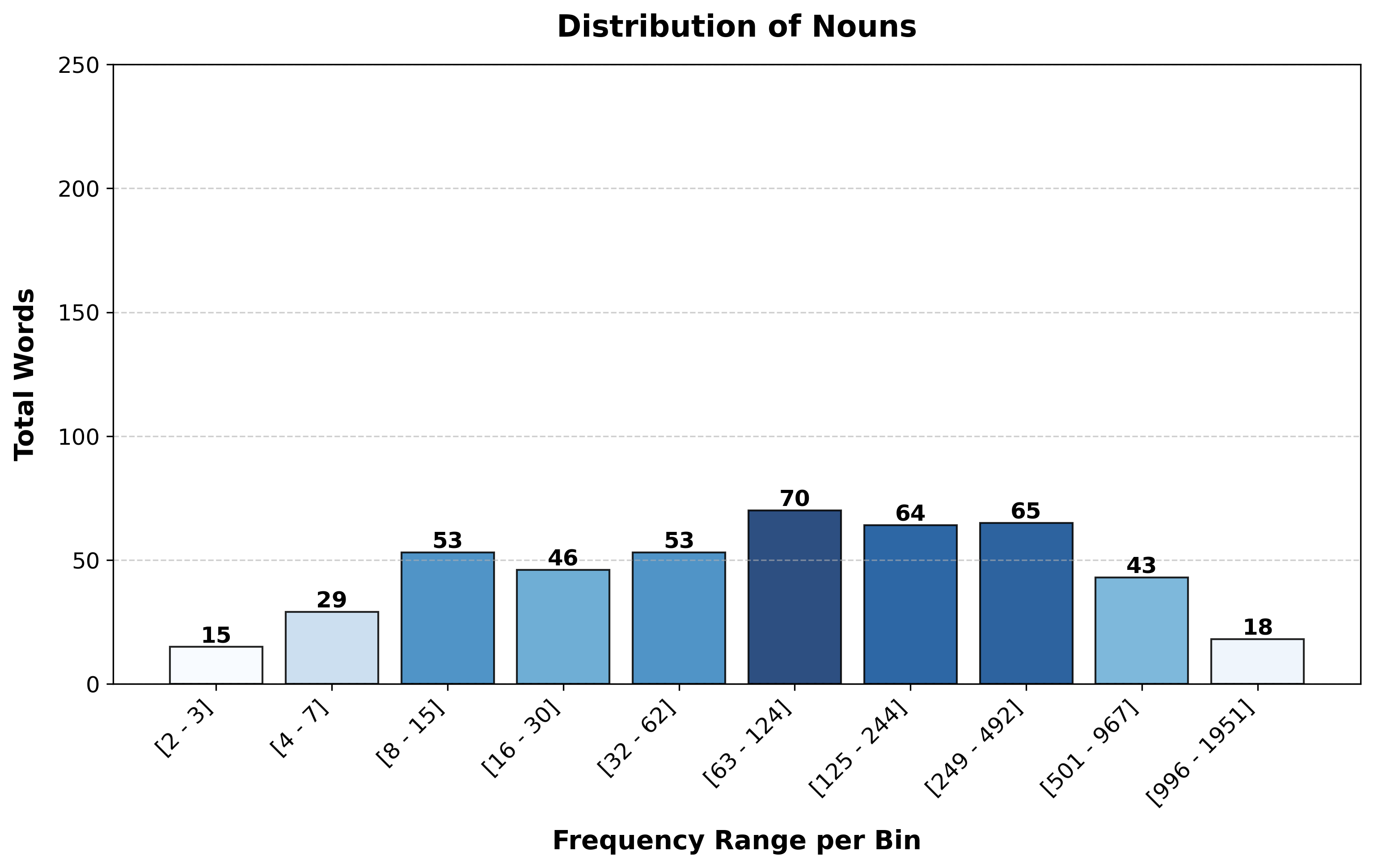}
        \caption{FR - Wiki}
    \end{subfigure}
    \hfill
    \begin{subfigure}[t]{0.22\textwidth}
        \centering
        \includegraphics[width=0.9\linewidth]{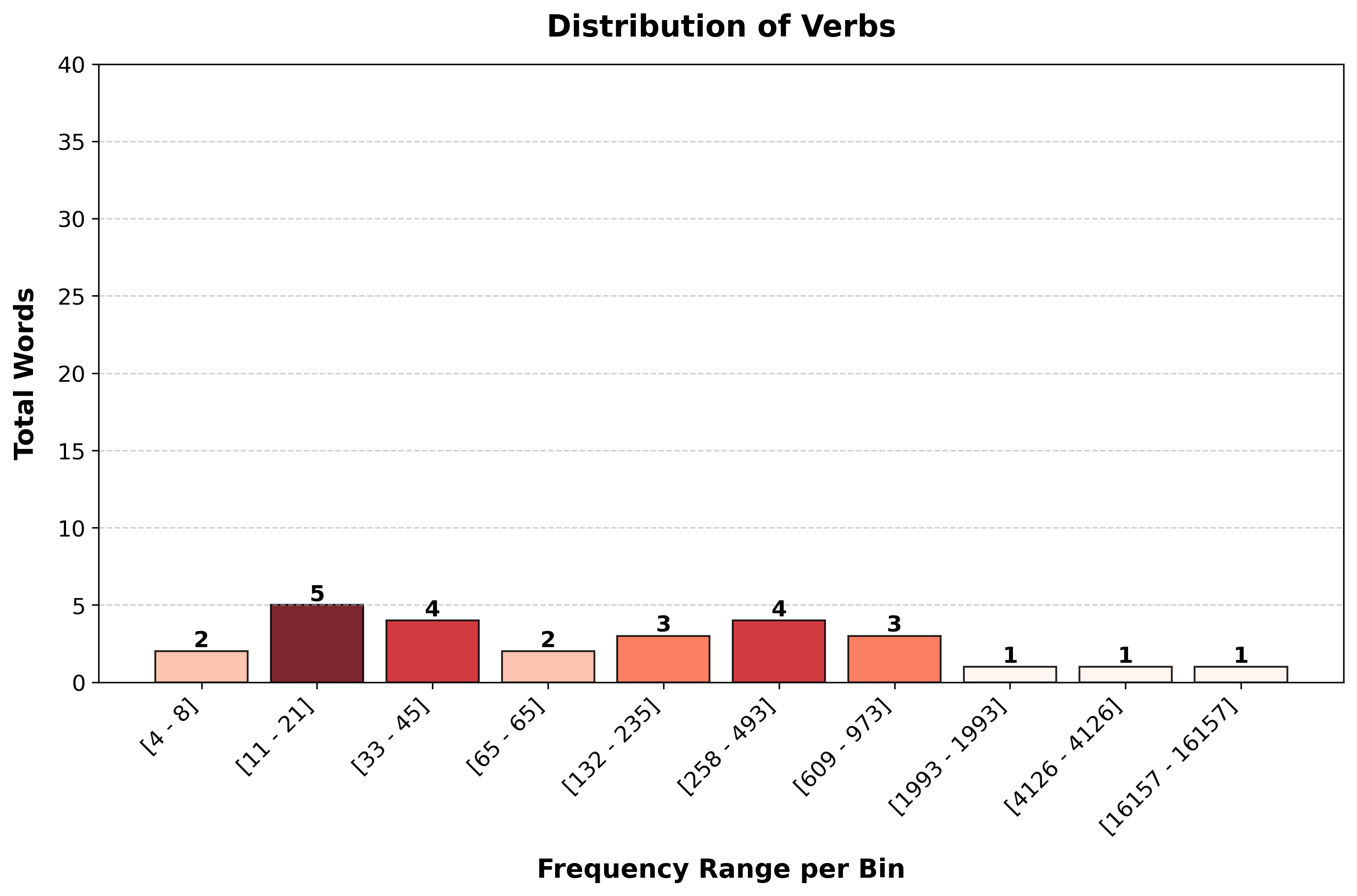}
        \caption{FR - CHILDES}
    \end{subfigure}
    \hfill
    \begin{subfigure}[t]{0.22\textwidth}
        \centering
        \includegraphics[width=0.9\linewidth]{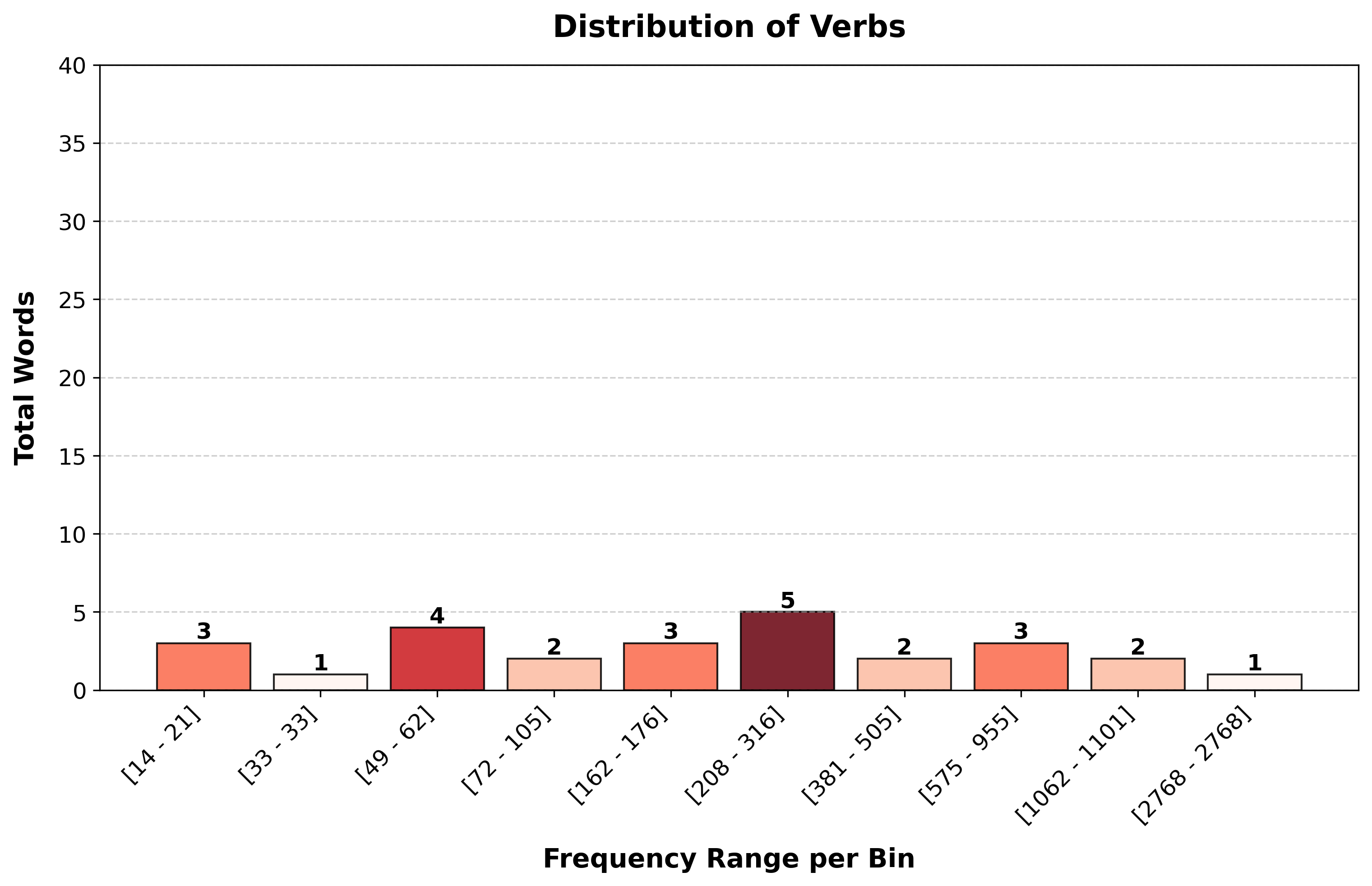}
        \caption{FR - Wiki}
    \end{subfigure}

    \vspace{1em}

    % Row 3
    \begin{subfigure}[t]{0.22\textwidth}
        \centering
        \includegraphics[width=0.9\linewidth]{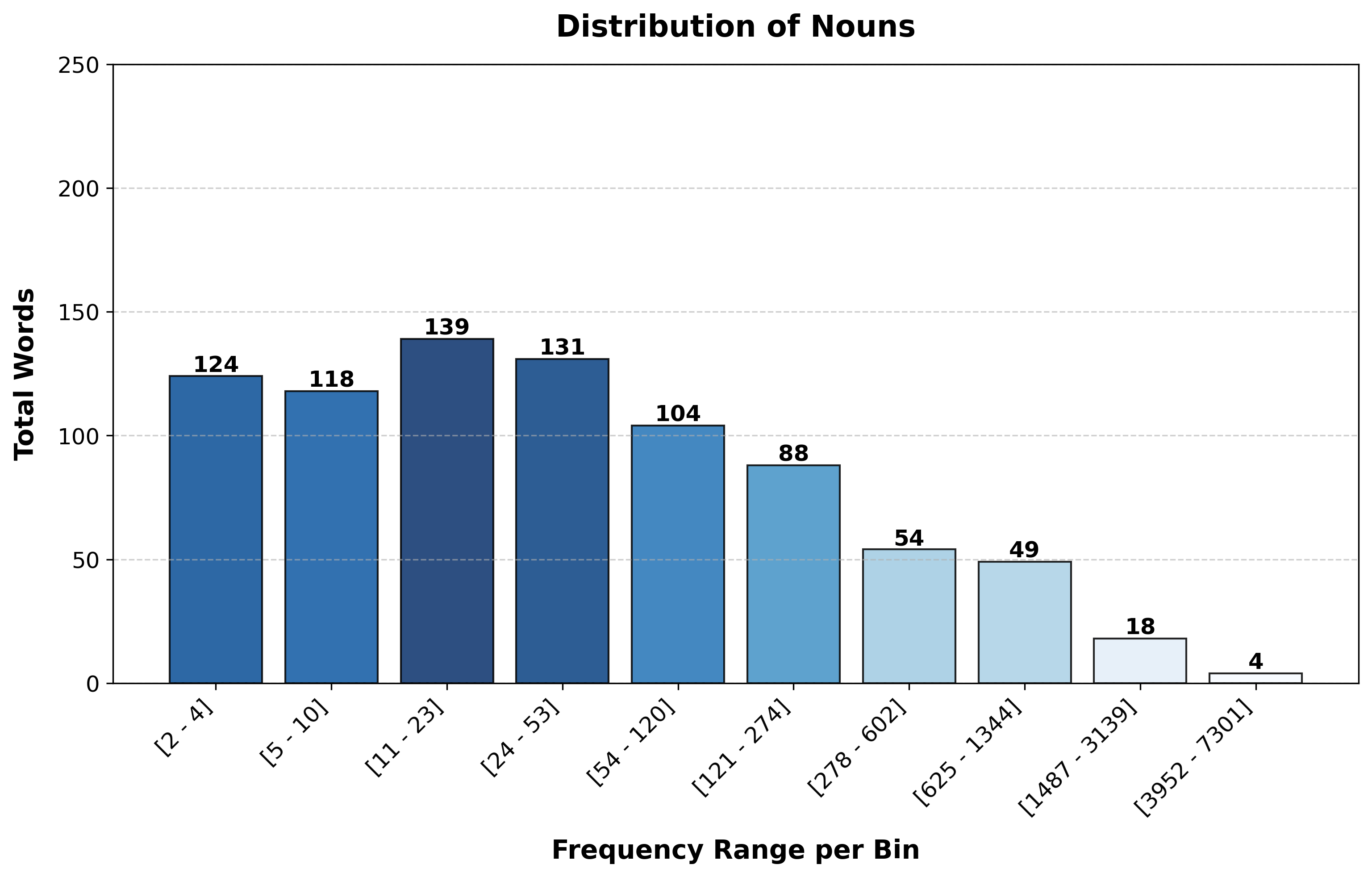}
        \caption{DE - CHILDES}
    \end{subfigure}
    \hfill
    \begin{subfigure}[t]{0.22\textwidth}
        \centering
        \includegraphics[width=0.9\linewidth]{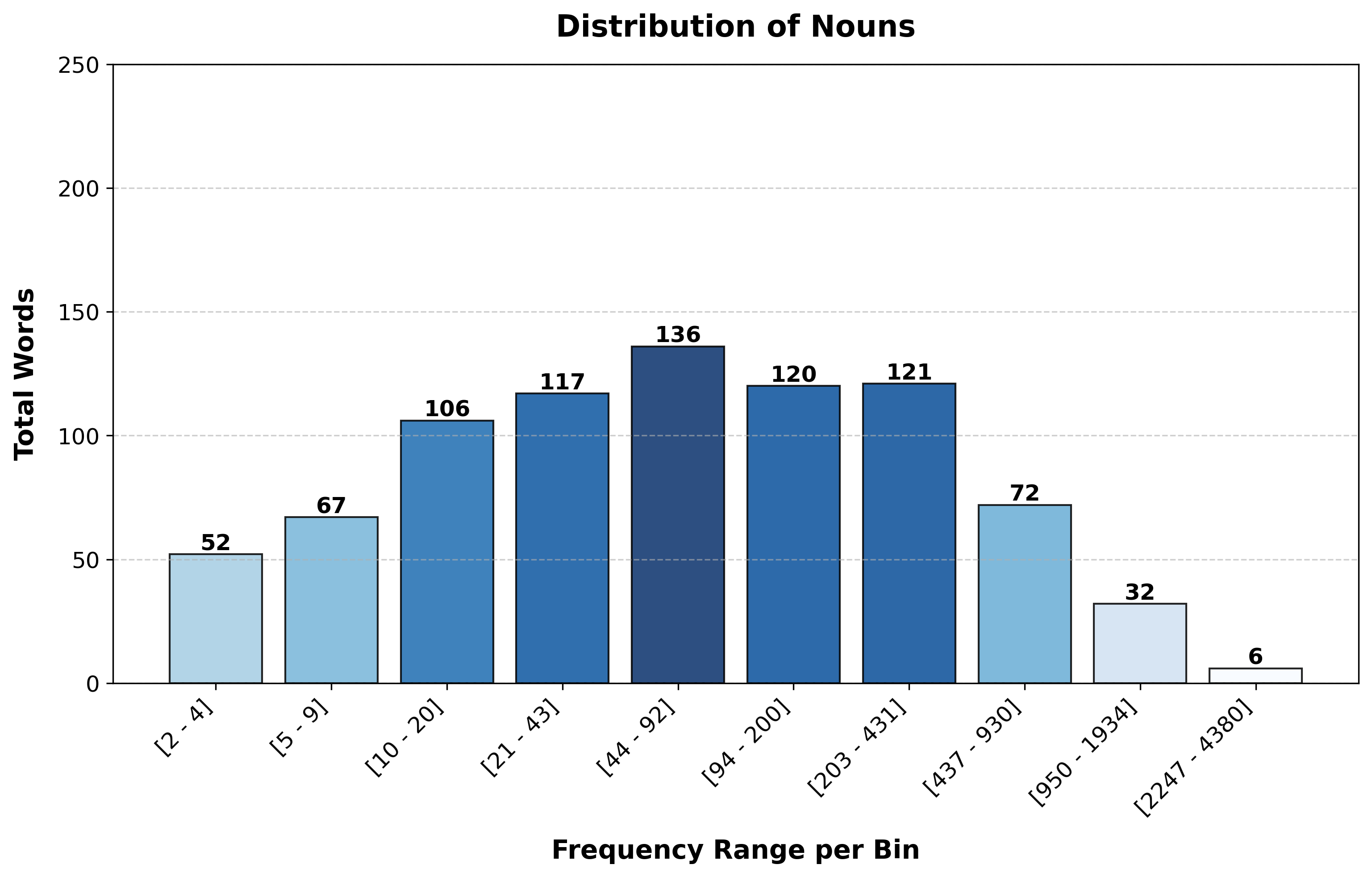}
        \caption{DE - Wiki}
    \end{subfigure}
    \hfill
    \begin{subfigure}[t]{0.22\textwidth}
        \centering
        \includegraphics[width=0.9\linewidth]{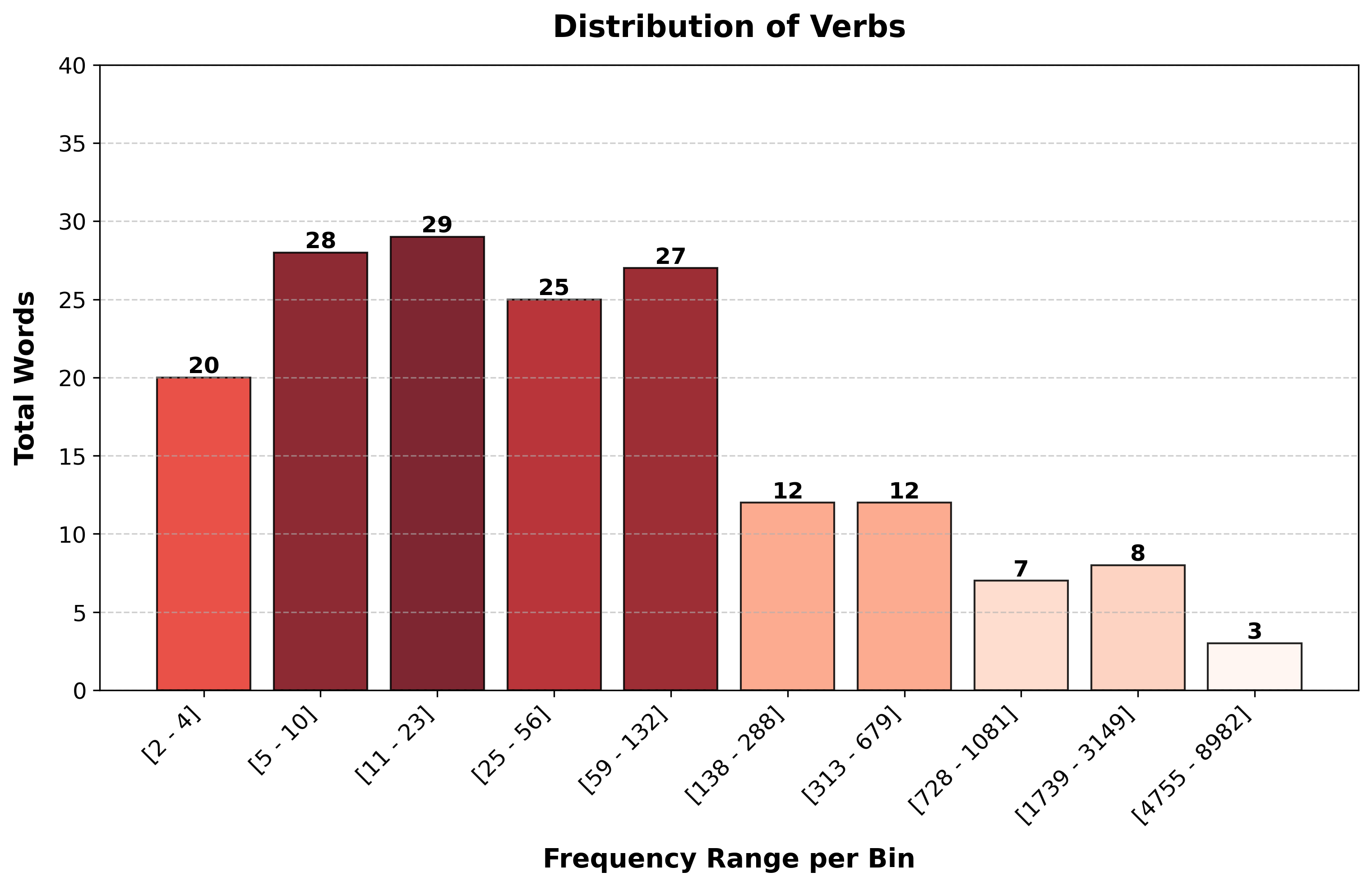}
        \caption{DE - CHILDES}
    \end{subfigure}
    \hfill
    \begin{subfigure}[t]{0.22\textwidth}
        \centering
        \includegraphics[width=0.9\linewidth]{imgs/DISTRIBUTION_VERBS/de/wiki_verbs_rerun_histogram_rerun.png}
        \caption{DE - Wiki}
    \end{subfigure}

    \caption{Noun and Verb Distribution across bins, in the two datasets and in the three languages.}
    \label{fig:three_rows_four_cols}
\end{figure*}

Curation pipeline details:

\begin{itemize}
    \item The shared vocabularies extracted for lexical selection include 14,809 tokens in English, 9,165 in French, and 19,187 in German. Frequency distributions are calculated using SpaCy’s \texttt{en\_core\_web\_sm}, \texttt{fr\_core\_web\_sm}, and \texttt{de\_core\_web\_sm} pipelines. For noun frequency analysis, singular and plural forms are counted separately. In German, case information is explicitly used to retain only nouns marked as nominative. For verbs, English selection includes forms tagged as \texttt{VB}, \texttt{VBP}, and \texttt{VBZ}, recognizing that these categories respectively capture infinitives, non-third-person present forms, and third-person singular present forms. In contrast, French and German verb selection involves additional morphological constraints: verbs must be third person and present tense, and forms in the subjunctive or conditional moods are excluded.

    \item As mentioned in the main body, the binning of candidate nouns and verbs into ten frequency categories was performed using a logarithmic scale. The bin edges were defined using log-spaced intervals between the minimum and maximum frequencies observed across the dataset. To visualize the distribution of nouns and verbs across the bins, reference can be made to the histograms in Figure \ref{fig:three_rows_four_cols}.

    \item For English and German, a total of 10 nouns and 10 verbs per dataset are retained. For French, due to constraints in the shared vocabulary, the final selection includes 9 nouns and 7 verbs. Additionally, two extra nouns per dataset are selected to serve as objects in the relative clause paradigms. The complete lists of selected lexical items for each language, along with their corresponding frequencies, are reported in Table~\ref{fig:nouns_verbs_selected}. It is also important to note that the relative clause paradigms include a verb within the relative clause itself. These verbs are adapted from CLAMS, with exclusions made for items not present in the shared vocabulary between CHILDES and Wikipedia. The final set of relative clause verbs used in our study is provided in Table~\ref{tab:rel_clauses}. Furthermore, the prepositions used in prepositional phrases are also drawn from CLAMS (~\Cref{tab:prepositions}). For paradigms involving long-distance dependencies within verb phrase coordination, the CLAMS minimal pairs include attractor nouns following both verbs in the coordinated structure. In our adaptation, we manually construct semantically appropriate fillers for each verb, without explicitly controlling for the frequency of the inserted lexical items.

\begin{table*}[htbp]
    \centering
    \scriptsize
    \caption{Selected Nouns (used as Subjects) and Verbs in the three languages from CHILDES and Wikipedia Distributions.}

    \begin{tabular}{@{}ll@{}}

    % Row 1: English (Nouns in CDL, Nouns in Wikipedia)
    % First minipage (English Nouns in CDL)
    \begin{minipage}{0.45\linewidth}
    \raggedright
    \tiny
    \begin{tabular}{llll}
    \toprule
    \textbf{EN Nouns}~ & \textbf{Bin} & \textbf{Freq} & \textbf{Df} \\
    \midrule
    roommate, roommates & 0 & 2 & CHI \\
    resident, residents & 1 & 6 & CHI \\
    librarian, librarians & 2 & 13 & CHI \\
    officer, officers & 3 & 36 & CHI \\
    toddler, toddlers & 4 & 90 & CHI \\
    farmer, farmers & 5 & 264 & CHI \\
    policeman, policemen & 6 & 380 & CHI \\
    doctor, doctors & 7 & 656 & CHI \\
    man, men & 8 & 2156 & CHI \\
    daddy, daddies & 9 & 7027 & CHI \\
    \bottomrule
    \end{tabular}
\end{minipage} &
\vspace{1em}

\begin{minipage}{0.45\linewidth}
    \tiny
    \raggedright
    \begin{tabular}{llll}
    \toprule
    \textbf{EN Nouns}~ & \textbf{Bin} & \textbf{Freq} & \textbf{Df} \\
    \midrule
    picker, pickers & 0 & 2 & Wiki \\
    harvester, harvesters & 0 & 3 & Wiki \\
    fireman, firemen & 1 & 11 & Wiki \\
    superhero, superheroes & 3 & 27 & Wiki \\
    explorer, explorers & 4 & 72 & Wiki \\
    painter, painters & 5 & 161 & Wiki \\
    parent, parents & 6 & 358 & Wiki \\
    writer, writers & 7 & 629 & Wiki \\
    president, presidents & 8 & 1473 & Wiki \\
    group, groups & 9 & 3085 & Wiki \\
    \bottomrule
    \end{tabular}
\end{minipage} \\

    % Vertical space between rows

    \begin{minipage}{0.45\linewidth}
    \raggedright
    \tiny
    \begin{tabular}{l l l l l}
    \toprule
    \textbf{EN Verbs}~ & \textbf{Bin} & \textbf{Freq} & \textbf{Long VP} & \textbf{Df} \\
    \midrule
    awaits, await & 0 & 2 & \textit{the guests} & CHI \\
    complains, complain & 1 & 8 & \textit{about the noise} & CHI \\
    arrives, arrive & 2 & 17 & \textit{at the station} & CHI \\
    disappears, disappear & 2 & 42 & \textit{from the scene} & CHI \\
    bows, bow & 4 & 243 & \textit{to the king} & CHI \\
    hides, hide & 4 & 391 & \textit{from the chicken} & CHI \\
    leaves, leave & 6 & 1793 & \textit{the room} & CHI \\
    sits, sit & 7 & 4219 & \textit{in the car} & CHI \\
    thinks, think & 8 & 14710 & \textit{about the trip} & CHI \\
    goes, go & 9 & 27620 & \textit{to the new store} & CHI \\
    \bottomrule
    \end{tabular}
\end{minipage} &
\vspace{1em}

% Second minipage (English Verbs in Wikipedia)
\begin{minipage}{0.45\linewidth}
    \raggedright
    \tiny
    \begin{tabular}{l l l l l}
    \toprule
    \textbf{EN Verbs}~ & \textbf{Bin} & \textbf{Freq} & \textbf{Long VP} & \textbf{Df} \\
    \midrule
    grinds, grind & 0 & 4 & \textit{the coffee beans} & Wiki \\
    exaggerates, exaggerate & 1 & 6 & \textit{with laughs} & Wiki \\
    screams, scream & 2 & 13 & \textit{very loudly} & Wiki \\
    swims, swim & 3 & 31 & \textit{in the pool} & Wiki \\
    enjoys, enjoy & 4 & 93 & \textit{the company of friends} & Wiki \\
    draws, draw & 5 & 212 & \textit{a nice picture} & Wiki \\
    rests, rest & 6 & 516 & \textit{on the couch} & Wiki \\
    runs, run & 6 & 975 & \textit{at the park} & Wiki \\
    plays, play & 7 & 1233 & \textit{with the toys} & Wiki \\
    works, work & 8 & 3545 & \textit{on a new project} & Wiki \\
    \bottomrule
    \end{tabular}
\end{minipage} \\

    % Vertical space between rows
    \vspace{1em}

   \begin{minipage}{0.45\linewidth}
    \raggedright
    \tiny
    \begin{tabular}{llll}
    \toprule
    \textbf{FR Nouns}~ & \textbf{Bin} & \textbf{Freq} & \textbf{Df} \\
    \midrule
    visiteur, visiteurs & 0 & 3 & CHI \\
    joueur, joueurs & 1 & 8 & CHI \\
    chanteur, chanteurs & 2 & 13 & CHI \\
    capitaine, capitaines & 3 & 32 & CHI \\
    homme, hommes & 5 & 84 & CHI \\
    pompier, pompiers & 6 & 171 & CHI \\
    dame, dames & 6 & 311 & CHI \\
    enfant, enfants & 7 & 667 & CHI \\
    lapin, lapins & 8 & 972 & CHI \\
    \bottomrule
    \end{tabular}
\end{minipage} &

% Second minipage (French Nouns in Wikipedia)
\begin{minipage}{0.45\linewidth}
    \raggedright
    \tiny
    \begin{tabular}{llll}
    \toprule
    \textbf{FR Nouns}~ & \textbf{Bin} & \textbf{Freq} & \textbf{Df}\\
    \midrule
    gamin, gamins & 0 & 3 & Wiki \\
    cuisinier, cuisiniers & 2 & 11 & Wiki \\
    vilaine, vilaines & 3 & 18 & Wiki \\
    avocat, avocats & 4 & 55 & Wiki \\
    pilote, pilotes & 6 & 192 & Wiki \\
    lecteur, lecteurs & 6 & 144 & Wiki \\
    prince, princes & 7 & 480 & Wiki \\
    personnage, personnages & 8 & 996 & Wiki \\
    groupe, groupes & 9 & 1610 & Wiki \\
    \bottomrule
    \end{tabular}
\end{minipage} \\

    % Vertical space between rows
    \vspace{1em}

    % Row 4: French (Verbs in CDL, Verbs in Wikipedia)
\begin{minipage}{0.45\linewidth}
    \raggedright
    \tiny
    \begin{tabular}{l l l l l}
    \toprule
    \textbf{FR Verbs}~ & \textbf{Bin} & \textbf{Freq} & \textbf{Long VP} & \textbf{Df} \\
    \midrule
    poursuit, poursuivent & 0 & 4 & \textit{une nouvelle mission} & CHI \\
    grandit, grandissent & 1 & 19 & \textit{très rapidement} & CHI \\
    apprend, apprennent & 3 & 65 & \textit{une nouvelle histoire} & CHI \\
    descend, descendent & 4 & 185 & \textit{les escaliers de la maison} & CHI \\
    attend, attendent & 5 & 258 & \textit{le repas chaud} & CHI \\
    arrive, arrivent & 6 & 973 & \textit{au lieu de rendez-vous} & CHI \\
    met, mettent & 7 & 1993 & \textit{la nappe sur la table} & CHI \\
    \bottomrule
    \end{tabular}
\end{minipage} &

% Second minipage (French Verbs in Wikipedia)
\begin{minipage}{0.45\linewidth}
    \raggedright
    \tiny
    \begin{tabular}{l l l l l}
    \toprule
    \textbf{FR Verbs}~ & \textbf{Bin} & \textbf{Freq} & \textbf{Long VP} & \textbf{Df} \\
    \midrule
    casse, cassent & 1 & 21 & \textit{le verre} & Wiki \\
    rentre, rentrent & 2 & 62 & \textit{dans la chambre} & Wiki \\
    continue, continuent & 5 & 223 & \textit{sur la route} & Wiki \\
    suit, suivent & 5 & 316 & \textit{le long chemin} & Wiki \\
    rend, rendent & 6 & 381 & \textit{le stylo à sa maman} & Wiki \\
    va, vont & 7 & 575 & \textit{au marché} & Wiki \\
    permet, permettent & 8 & 1062 & \textit{l'accès aux escaliers} & Wiki \\
    \bottomrule
    \end{tabular}
\end{minipage} \\

    % Vertical space between rows
    \vspace{1em}

    \begin{minipage}{0.45\linewidth}
    \raggedright
    \tiny
    \begin{tabular}{llll}
    \toprule
    \textbf{DE Nouns}~ & \textbf{Bin} & \textbf{Freq} & \textbf{Df} \\
    \midrule
    feind, feinde & 0 & 4 & CHI \\
    architekt, architekten & 0 & 4 & CHI \\
    präsident, präsidenten & 1 & 6 & CHI \\
    kollege, kollegen & 2 & 17 & CHI \\
    ingenieur, ingenieure & 3 & 26 & CHI \\
    sohn, söhne & 4 & 96 & CHI \\
    arzt, ärzte & 5 & 161 & CHI \\
    doktor, doktoren & 6 & 295 & CHI \\
    mensch, menschen & 7 & 1247 & CHI \\
    frau, frauen & 8 & 1841 & CHI \\
    \bottomrule
    \end{tabular}
\end{minipage} &

% Second minipage (German Nouns in Wikipedia)
\begin{minipage}{0.45\linewidth}
    \raggedright
    \tiny
    \begin{tabular}{l l l l}
    \toprule
    \textbf{DE Nouns}~ & \textbf{Bin} & \textbf{Freq} & \textbf{Df}\\
    \midrule
    fahrgast, fahrgäste & 1 & 8 & Wiki \\
    kleinkind, kleinkinder & 2 & 12 & Wiki \\
    zwilling, zwillinge & 3 & 23 & Wiki \\
    polizist, polizisten & 3 & 39 & Wiki \\
    kunde, kunden & 5 & 105 & Wiki \\
    schwester, schwestern & 5 & 171 & Wiki \\
    bruder, brüder & 6 & 374 & Wiki \\
    vater, väter & 7 & 736 & Wiki \\
    mann, männer & 7 & 642 & Wiki \\
    person, personen & 8 & 1114 & Wiki \\
    \bottomrule
    \end{tabular}
\end{minipage} \\

    % Vertical space between rows
    \vspace{1em}

    % Row 6: German (Verbs in CDL, Verbs in Wikipedia)
    \begin{minipage}{0.45\linewidth}
    \raggedright
    \tiny
    \begin{tabular}{l l l l l}
    \toprule
    \textbf{DE Verbs}~ & \textbf{Bin} & \textbf{Freq} & \textbf{Long VP} & \textbf{Df} \\
    \midrule
    zweifelt, zweifeln & 0 & 4 & \textit{am wetter} & CHI \\
    konstruiert, konstruieren & 1 & 5 & \textit{ein modell} & CHI \\
    fürchtet, fürchten & 3 & 31 & \textit{den starken sturm} & CHI \\
    schält, schälen & 3 & 40 & \textit{den reifen grünen apfel} & CHI \\
    taucht, tauchen & 4 & 64 & \textit{in das wasser des meeres} & CHI \\
    kennt, kennen & 5 & 259 & \textit{die antwort auf die frage} & CHI \\
    schreibt, schreiben & 7 & 865 & \textit{einen brief an verwandte} & CHI \\
    erzählt, erzählen & 7 & 1081 & \textit{eine geschichte über die ferien} & CHI \\
    spielt, spielen & 8 & 3149 & \textit{mit dem ball auf dem hof} & CHI \\
    kommt, kommen & 9 & 8982 & \textit{mit dem bus zum tennisplatz} & CHI \\
    \bottomrule
    \end{tabular}
\end{minipage} &

% Second minipage (German Verbs in Wikipedia)
\begin{minipage}{0.45\linewidth}
    \raggedright
    \tiny
    \begin{tabular}{l l l l l}
    \toprule
    \textbf{DE Verbs}~ & \textbf{Bin} & \textbf{Freq} & \textbf{Long VP} & \textbf{Df} \\
    \midrule
    schaukelt, schaukeln & 0 & 2 & \textit{auf dem spielplatz} & Wiki \\
    flüchtet, flüchten & 2 & 13 & \textit{vor dem feuer} & Wiki \\
    riecht, riechen & 2 & 12 & \textit{den duft von frischem kaffee} & Wiki \\
    wandert, wandern & 4 & 36 & \textit{durch den wald} & Wiki \\
    feiert, feiern & 4 & 48 & \textit{den geburtstag des großvaters} & Wiki \\
    verschwindet, verschwinden & 5 & 68 & \textit{im nebel} & Wiki \\
    denkt, denken & 6 & 210 & \textit{an blumen im garten} & Wiki \\
    spricht, sprechen & 7 & 529 & \textit{über das abendessen} & Wiki \\
    arbeitet, arbeiten & 8 & 640 & \textit{an einem projekt} & Wiki \\
    liegt, liegen & 9 & 2220 & \textit{auf dem boden} & Wiki \\
    \bottomrule
    \end{tabular}
\end{minipage} \\
    \end{tabular}
    \label{fig:nouns_verbs_selected}
\end{table*}

\begin{table*}[htbp]
\caption{Chosen Nouns (used as Objects) in FIT-CLAMS for English, French and German.}
\centering
\begin{minipage}[t]{0.32\textwidth}
\centering
\tiny
\vspace{0.2em}
\begin{tabular}{lllll}
\toprule
\textbf{English Nouns}  & \textbf{Bin} & \textbf{Freq}  & \textbf{Df} \\
\midrule
guard, guards & 3 & 35 & CHILDES \\
friend, friends & 7 & 1414 & CHILDES \\
waiter, waiters & 1 & 10 & Wiki \\
speaker, speakers & 6 & 347 & Wiki \\
\bottomrule
\end{tabular}
\end{minipage}
\hfill
\begin{minipage}[t]{0.32\textwidth}
\centering
\tiny
\vspace{0.2em}
\begin{tabular}{llll}
\toprule
\textbf{French Nouns} & \textbf{Bin} & \textbf{Freq}  & \textbf{Df} \\
\midrule
femme, femmes  & 4 & 71 & CHILDES \\
adulte, adultes  & 3 & 35 & CHILDES \\
constructeur, constructeurs & 5 & 106 &  Wiki \\
docteur, docteurs & 5 & 85  & Wiki \\
\bottomrule
\end{tabular}
\end{minipage}
\hfill
\begin{minipage}[t]{0.32\textwidth}
\centering
\tiny
\vspace{0.2em}
\begin{tabular}{llll}
\toprule
\textbf{German Nouns} & \textbf{Bin} & \textbf{Freq} & \textbf{Df} \\
\midrule
mitglied, mitglieder  & 1 & 8 & CHILDES \\
bauer, bauern   & 6 & 332 & CHILDES \\
matrose, matrosen & 1 & 11  & Wiki \\
familie, familien  & 8 & 959  & Wiki \\
\bottomrule
\end{tabular}
\end{minipage}

\end{table*}
\label{tab:object_selected}

\begin{table*}[htbp]
\centering
\small
\begin{tabular}{ll}
\toprule
\textbf{Language} & \textbf{Prepositions} \\
\midrule
\textbf{English} & next to, behind, in front of, near, to the side of, across from \\
\textbf{French}  & devant, derrière, en face, à côté, près \\
\textbf{German}  & vor, hinter, neben, in der Nähe von, gegenüber \\
\bottomrule
\end{tabular}
\caption{Prepositions used in FIT-CLAMS for English, French, and German.}
\label{tab:prepositions}
\end{table*}

\begin{table*}[htbp]
\centering
\small
\begin{tabular}{ll}
\toprule
\textbf{Language} & \textbf{Verbs Used in Relative Clauses} \\
\midrule
\textbf{English} & like likes; hates hate; love loves; admires admire \\
\textbf{French}  & aime aiment \\
\textbf{German}  & mag mögen; vermeidet vermeiden \\
\bottomrule
\end{tabular}
\caption{Verbs used in FIT-CLAMS relative clauses for English, French, and German.}
\label{tab:rel_clauses}
\end{table*}

\begin{table*}[ht]
\centering
\footnotesize
\begin{tabular}{lccc}
\toprule
\textbf{Paradigm} & \textbf{EN} & \textbf{FR} & \textbf{DE} \\
\midrule
Agreement in long VP coordinates & 900 & 378 & 900\\
Agreement in object relative clauses (across) &3200 & 504 & 1600 \\
Agreement in object relative clauses (within) &3200 & 504 & 1600 \\
Agreement in prep phrases & 4800 & 2520 & 4000 \\
Simple agreement & 200 & 126 & 200\\
Agreement in subject relative clauses &3200 & 504 & 1600 \\
Agreement in VP coordinates & 900 & 378 & 900\\
\bottomrule
\end{tabular}
\caption{Minimal pair counts of FIT-CLAMS (same for FIT-CLAMS-C and FIT-CLAMS-W) for each paradigm across three languages.}
\label{tab:minimal-pairs}
\end{table*}

\end{itemize}

All generated sentences in English and French have been manually reviewed by the authors, who have linguistic expertise in these two languages, while the German sentences were validated by a native speaker to ensure grammaticality and naturalness.

\label{sec:noun_verb_distrib_bins}

\end{document}